\documentclass{article}

    \PassOptionsToPackage{numbers, compress}{natbib}
\usepackage[preprint]{neurips_2023}

\usepackage[utf8]{inputenc} % allow utf-8 input
\usepackage[T1]{fontenc}    % use 8-bit T1 fonts
\usepackage[pagebackref,breaklinks,citecolor=citecolor,colorlinks]{hyperref}       % hyperlinks
\usepackage{url}            % simple URL typesetting
\usepackage{booktabs}       % professional-quality tables
\usepackage{amsfonts}       % blackboard math symbols
\usepackage{nicefrac}       % compact symbols for 1/2, etc.
\usepackage{microtype}      % microtypography
\usepackage{xcolor}         % colors
\usepackage{graphicx}
\usepackage{amsmath}
\usepackage{amssymb}
\usepackage{color}
\usepackage{ctable}
\usepackage{colortbl}
\usepackage{lipsum}
\usepackage{multirow}
\usepackage{soul}
\usepackage{xspace}\xspace
\usepackage{adjustbox}
\usepackage{array}
\usepackage{epsfig}
\usepackage{bbding}
\usepackage{dblfloatfix}
\usepackage{transparent}
\usepackage{diagbox}
\usepackage{listings}
\usepackage{times}
\usepackage{bm}
\usepackage{bbm}
\usepackage{amsbsy}
\usepackage{cases}
\usepackage{enumitem}
\usepackage{subcaption}
\usepackage[titletoc]{appendix}
\usepackage{grffile}
\usepackage{pifont}
\usepackage{multicol}
\usepackage{comment}
\definecolor{citecolor}{HTML}{0071bc}

\usepackage{diagbox}
\usepackage{etoc}
\etocdepthtag.toc{mtchapter}
\etocsettagdepth{mtchapter}{subsection}
\etocsettagdepth{mtappendix}{none}

\newcommand{\ourMethod}{DiffComplete}

\newcommand{\ie}{\textit{i.e.}}
\newcommand{\eg}{\textit{e.g.}}

\title{DiffComplete: Diffusion-based Generative\\ 3D Shape Completion}

\author{Ruihang Chu$^{1}$\hspace{1.5cm}Enze Xie $^{2}$\hspace{1.5cm}Shentong Mo$^{3}$\hspace{1.5cm}\\
\textbf{Zhenguo Li$^{2}$}\hspace{0.6cm}\textbf{Matthias Nießner$^{4}$}\hspace{0.6cm}\textbf{Chi-Wing Fu$^{1}$}\hspace{0.6cm}\textbf{Jiaya Jia$^{1}$}\\
$^{1}$CUHK~~~~~
$^{2}$Huawei Noah's Ark Lab~~~~~
$^{3}$MBZUAI~~~~~
$^{4}$TUM\\
{\small \url{https://ruihangchu.com/diffcomplete.html}}
}

\begin{document}

\maketitle

\begin{abstract}
We introduce a new diffusion-based approach for shape completion on 3D range scans.
Compared with prior deterministic and probabilistic methods, we strike a balance between realism, multi-modality, and high fidelity.
We propose \ourMethod~by casting shape completion as a generative task conditioned on the incomplete shape.
Our key designs are two-fold.
First, we devise a hierarchical feature aggregation mechanism to inject conditional features in a spatially-consistent manner.
So, we can capture both local details and broader contexts of the conditional inputs to control the shape completion.
Second, we propose an occupancy-aware fusion strategy in our model to enable the completion of multiple partial shapes and introduce higher flexibility on the input conditions.
\ourMethod~sets a new SOTA performance (\eg, 40\% decrease on $l_1$ error) on two large-scale 3D shape completion benchmarks.
Our completed shapes not only have a realistic outlook compared with the deterministic methods but also exhibit high similarity to the ground truths compared with the probabilistic alternatives.
Further, \ourMethod~has strong generalizability on objects of entirely unseen classes for both synthetic and real data, eliminating the need for model re-training in various applications.

\end{abstract}

\begin{figure}[htbp]
	\begin{center}
  \vspace*{-2mm}
  \includegraphics[width=0.9\columnwidth]{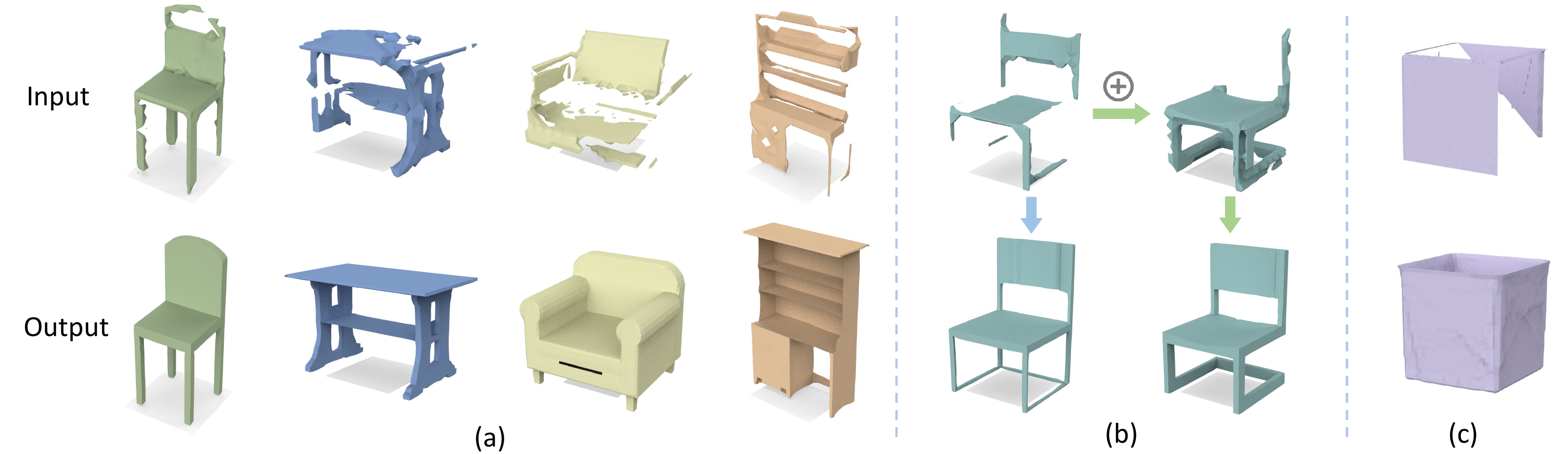}
	\end{center}
	\vspace*{-3mm}
	\caption{Our method is able to (a) produce realistic completed shapes from partial scans, (b) incorporate multiple incomplete scans (denoted by the plus symbol) to improve the completion accuracy, and (c) directly generalize to work on real objects of unseen classes without finetuning.}
	\label{fig:teaser}
\vspace*{-3mm}
\end{figure}

\section{Introduction}

The advent of affordable range sensors, such as the Microsoft Kinect and Intel RealSense, has spurred remarkable progress in 3D reconstruction~\cite{reco_star}, facilitating applications in content creation, mixed reality, robot navigation, and more.
Despite the improved reconstruction quality~\cite{dai2017bundlefusion,azinovic2022neural,park2019deepsdf,yariv2020multiview}, typical scanning sessions cannot cover every aspect of the scene objects.
Hence, the reconstructed 3D models often have incomplete geometry, thereby limiting their practical usage in downstream applications.
To fully unlock the potential of 3D reconstruction for supporting assorted applications, it is essential to address the challenges of completing incomplete shapes.

Effectively, a shape completion should produce shapes that are \textit{realistic}, \textit{probabilistic}, and \textit{high-fidelity}.
First, the produced shapes should look plausible without visual artifacts.
Second, considering the under-determined nature of completion, it is desirable for the model to generate diverse, multimodal outputs when filling in the missing regions, to improve the likelihood of obtaining a superior shape and enable creative use. 
Third, while multiple outputs encourage the model's generative ability, maintaining effective control over the completion results is crucial to ensure coherent reconstructions that closely resemble the ground-truth shapes.

Current approaches to 3D shape completion fall into deterministic and probabilistic paradigms. The former class excels at aligning predictions with ground truths, owing to their determined mapping functions and full supervision. However, this can expose the models to a higher risk of over-fitting, leading to undesirable artifacts, such as rugged surfaces, especially on
unseen objects. 
On the other hand, probabilistic approaches formulate the shape completion as a conditional generation task to produce plausible results, paired with techniques like auto-encoding~\cite{mittal2022autosdf}, adversarial training~\cite{wu2020multimodal,smith2017improved,zhang2021unsupervised}, or diffusion models~\cite{muller2022diffrf,cheng2022sdfusion,zhou20213d,chou2022diffusionsdf}.
These approaches mainly focus on cases that prioritize completion diversity, for instance, filling in large missing regions cropped out by 3D boxes, where the algorithm has more freedom to explore various plausible shapes.
However, when completing shapes obtained from range scans/reconstructions, whose geometries are often contaminated with noise, having varying degrees of incompleteness (see row 1 in Fig.~\ref{fig:teaser}), the goal is to recover the ground-truth scanned objects as accurately as possible. Therefore, relying solely on previous probabilistic approaches may compromise the completion accuracy and leads to low-fidelity results.

In this work, we focus on completing shapes acquired from range scans, 
aiming to produce \textit{realistic} and \textit{high-fidelity} completed shapes, while considering also the \textit{probabilistic} uncertainty.
We approach this task as a conditional generation with the proposed diffusion model named~\ourMethod.
To address the accuracy challenge typically in probabilistic models, we introduce a key design that incorporates range scan features in a hierarchical and spatially-consistent manner to facilitate effective controls.
Additionally, we propose a novel occupancy-aware fusion strategy that
allows taking multiple partial shapes as input for more precise completions.

Specifically, we formulate the diffusion process on a truncated distance field in a volumetric space. This representation allows us to encode both complete and incomplete shapes into multi-resolution structured feature volumes, thereby enabling their interactions at various network levels. At each level, features are aggregated based on voxel-to-voxel correspondence, enabling precise correlation of the difference in partialness between the shapes. 
By leveraging multi-level aggregation, the completion process can be controlled by both local details and broader contexts of the conditional inputs to improve the completion accuracy and network generalizability, as shown in Sec.~\ref{subsec:ablation}.

To condition the completion on multiple partial shapes, our occupancy-aware fusion strategy adopts the occupancy mask as a weight to adaptively fuse all observed geometry features from various incomplete shapes, and then employ the fused features to control the completion process.
For robustness, such an operation is performed in a multi-scale feature space. With our novel design, switching between single and multiple conditions can be efficiently achieved by finetuning an MLP layer. As Fig.~\ref{fig:teaser}(b) and~\ref{fig:multi_input} show, using multiple conditions for completion progressively refines the final shapes towards the ground truths.

\ourMethod~sets a new state-of-the-art performance on two large-scale shape completion benchmarks~\cite{dai2017shape,rao2022patchcomplete} in terms of completion accuracy and visual quality.
Compared to prior deterministic and probabilistic approaches, we not only generate multimodal plausible shapes with minimal artifacts but also present high shape fidelity relative to the ground truths. 
Further, \ourMethod~can directly generalize to work on unseen object categories.
Without specific zero-shot designs, it surpasses all existing approaches for both synthetic and real-world data, allowing us to eliminate the need for model re-training in various applications. To sum up, our contributions are listed as follows:

\begin{itemize}
    \item We introduce a diffusion model to complete 3D shapes on range scans, enabling realistic, probabilistic, and high-fidelity 3D geometry.
    \item To improve the completion accuracy, we enable effective control from single-input condition by hierarchical and spatially-consistent feature aggregation, and from multiple ones by an occupancy-aware fusion strategy.
    \item We show SOTA performance on shape completion on both novel instances and entirely unseen object categories, along with an in-depth analysis on a range of model characteristics.
\end{itemize}

\section{Related Work}
\label{sec:related_work}

\noindent
\textbf{RGB-D reconstruction.}
Traditional methods rely on geometric approaches for 3D reconstruction~\cite{engel2014lsd,mur2015orb,whelan2015elasticfusion,maier2017intrinsic3d,zollhoefer2015shading}.
A pioneering method~\cite{DBLP:conf/siggraph/CurlessL96} proposes a volumetric fusion strategy to integrate multiple range images into a unified 3D model on truncated signed distance fields (TSDF), forming the basis for many modern techniques like KinectFusion~\cite{izadi2011kinectfusion,newcombe2011kinectfusion}, VoxelHashing~\cite{niessner2013real}, and BundleFusion~\cite{dai2017bundlefusion}. 
Recent learning-based approaches further improve the reconstruction quality with fewer artifacts~\cite{weder2020routedfusion,peng2020convolutional,li2020multi,dai2021spsg,azinovic2022neural}, yet the intrinsic occlusions and measurement noise of 3D scans constrain the completeness of 3D reconstructions, making them still less refined than manually-created assets.

\noindent
\textbf{3D shape completion}~is a common post-processing step to fill in missing parts in the reconstructed shapes.
Classic methods mainly handle small holes and geometry primitives via Laplacian hole filling~\cite{sorkine2004least,nealen2006laplacian,zhao2007robust} or Poisson surface reconstruction~\cite{kazhdan2006poisson,kazhdan2013screened}.
Another line exploits the structural regularities of 3D shapes, such as the symmetries, to predict the unobserved data~\cite{thrun2005shape,mitra2006partial,pauly2008discovering,sipiran2014approximate,speciale2016symmetry}.

The availability of large 3D data has sparked retrieval-based methods~\cite{sung2015data,li2015database,nan2012search,kim2012acquiring} and learning-based fitting methods~\cite{nguyen2016field,firman2016structured,dai2017shape,dai2020sg,yu2021pointr,han2017high,song2017semantic,chibane2020implicit}. The former retrieves the shapes from a database that best match the incomplete inputs, whereas the latter minimizes the difference between the network-predicted shapes and ground truths.
3D-EPN~\cite{dai2017shape}, for instance, proposes a 3D encoder-decoder architecture to predict the complete shape from partial volumetric data.
Scan2Mesh~\cite{dai2019scan2mesh} converts range scans into 3D meshes with a direct optimization on the mesh surface.
PatchComplete~\cite{rao2022patchcomplete} further leverages local structural priors for completing shapes of unseen categories.

Generative methods, e.g., GANs~\cite{zheng2022sdf,zhang2021unsupervised,chen2019unpaired,wu2020multimodal,smith2017improved} and AutoEncoders~\cite{mittal2022autosdf,achlioptas2018learning}, offer an alternative to shape completion. 
Being able to generate diverse plausible global shapes based on a partial input, they allow for high generation freedom and thus compromise the completion accuracy in scenarios where a ground truth exists~\cite{zheng2022sdf}. 
Distinctively, we employ a powerful diffusion model for shape completion, specifically designed to prioritize the fidelity relative to ground truths while retaining the output diversity.
Compared to both generative and fitting-based paradigms, our method also effectively reduces surface artifacts, producing more realistic and natural 3D shapes.
In addition, we show superior generalization ability on completing objects of novel classes over SOTAs.

\noindent
\textbf{Diffusion models for 3D generation.} Diffusion models~\cite{sohl2015deep,dhariwal2021diffusion,rombach2022high,song2020score,nichol2021improved,ho2020denoising,song2020denoising,sinha2021d2c} have shown superior performance in various generation tasks, outperforming GANs' sample quality while preserving the likelihood evaluation property of VAEs. When adopted in 3D domain, a range of works~\cite{zhou20213d,zeng2022lion,luo2021diffusion,nichol2022point} focus on point cloud generation. For more complex surface generation, some works~\cite{nam20223d,chou2022diffusionsdf,muller2022diffrf,zhang20233dshape2vecset,bautista2022gaudi,cheng2022sdfusion} adopt latent diffusion models to learn implicit neural representations and form final shapes via a decoder. 
While some others~\cite{muller2022diffrf,cheng2022sdfusion,zhou20213d,chou2022diffusionsdf} support conditional shape completion, they typically fill in large regions cropped out with 3D boxes or reconstruct reasonable shapes from images. Due to the absence of meaningful ground truths in these completion scenarios, they could also face completion accuracy challenges like the above generative approaches ~\cite{zheng2022sdf,zhang2021unsupervised,chen2019unpaired,wu2020multimodal,smith2017improved}.
Instead, our method concentrates on completing shapes acquired from range scans, yielding outputs closely resembling the actual objects scanned. 
Also, we apply the diffusion process in explicit volumetric TSDFs, preserving more geometrical structures during the completion process.

\section{Method}

\subsection{Formulation}
\label{subsec:formulation}
To prepare the training data, we generate an incomplete 3D scan from depth frames using volumetric fusion~\cite{DBLP:conf/siggraph/CurlessL96} and represent the scan as a truncated signed distance field (TSDF) in a volumetric grid.
Yet, to accurately represent a ground-truth shape, we choose to use a volumetric truncated unsigned distance field (TUDF) instead. This is because retrieving the sign bit from arbitrary 3D CAD models (\eg, some are not closed) is non-trivial. By using TUDF, we can robustly capture the geometric features of different objects without being limited by the topology.

Given such a volume pair,~\ie, the incomplete scan $c$ and complete 3D shape $x_0$, we formulate the shape completion as a generation task that produces $x_0$ conditioned on $c$. We employ the probabilistic diffusion model as our generative model. In particular, it contains (i) a \textit{forward process} (denoted as $q(x_{0:T})$) that gradually adds Gaussian noise to corrupt the ground-truth shape $x_0$ into a random noise volume $x_T$, where $T$ is the total number of time steps; and (ii) a \textit{backward process} that employs a shape completion network, with learned parameters $\theta$, to iteratively remove the noise from $x_T$ conditioned on the partial scan $c$ and produce the final shape, denoted as $p_\theta(x_{0:T}, c)$.
As both the $\textit{forward}$ and $\textit{backward}$ processes are governed by a discrete-time Markov chain with time steps $\{0, . . . , T \}$, their Gaussian transition probabilities can be formulated as
\begin{gather}
q(x_{0:T})=q(x_0)\prod_{t=1}^{T}q(x_t|x_{t-1}), \quad q(x_t|x_{t-1}):=\mathcal{N}(\sqrt{1-\beta_t}x_{t-1},\beta_t\mathbf{I})
\label{eq:forward}\\
\text{and} \ \ 
p_\theta(x_{0:T},c)=p(x_T)\prod_{t=1}^{T}p_\theta(x_{t-1}|x_t,c), \quad p_\theta(x_{t-1}|x_t):=\mathcal{N}(\mu_\theta(x_t,t,c), \sigma^2_t\mathbf{I}).
\label{eq:backward}
\end{gather}

In the \textit{forward} process (Eq.~\eqref{eq:forward}), the scalars $\beta_t\in[0,1]$ control a variance schedule that defines the amount of noise added in each step $t$.
In the \textit{backward} process (Eq.~\eqref{eq:backward}), $p(x_T)$ is a Gaussian prior in time step $t$, $\mu_\theta$ represents the mean predicted from our network and $\sigma_t^2$ is the variance.
As suggested in DDPM~\cite{ho2020denoising}, predicting $\mu_\theta(x_t,t,c)$ can be simplified to alternatively predicting $\epsilon_\theta(x_t,t,c)$, which should approximate the noise used to corrupt $x_{t-1}$ in the \textit{forward} process, and $\sigma_t$ can be replaced by the pre-defined $\beta_t$. 
With these modifications, we can optimize the network parameter $\theta$ with a mean squared error loss to maximize the generation probability $p_\theta(x_0)$.
The training objective is
\begin{equation}  \mathop{\arg\min}\limits_{\theta}E_{t,x_0,\epsilon,c}[||\epsilon-\epsilon_\theta(x_t,t,c)||^2],\quad\epsilon\in\mathcal{N}(0,\mathbf{I})
\label{eq:train_obj}
\end{equation}
where $\epsilon$ is the applied noise to corrupt $x_0$ into $x_t$ and $\mathcal{N}(0,\mathbf{I})$ denotes a unit Gaussian distribution.
We define all the diffusion processes in a volume space of a resolution $S^3$,~\ie, $x_{0:T},c,\epsilon\in\mathbb{R}^{S\times S\times S}$, where each voxel stores a scalar TSDF/TUDF value; $S$=32 in our experiments.
Compared to previous latent diffusion models~\cite{cheng2022sdfusion,chou2022diffusionsdf,nam20223d,muller2022diffrf} that require shape embedding first, we directly manipulate the shape with a better preservation of geometric features. Doing so naturally enables our hierarchical feature aggregation strategy (see Sec.~\ref{subsec:framework}).
Next, we will introduce how to predict $\epsilon_\theta(x_t,t,c)$.

\subsection{Shape Completion Pipeline}
\label{subsec:framework}

\textbf{Overview.} To enhance the completion accuracy, we encourage the incomplete scans to control completion behaviors.
Inspired from the recent ControlNet~\cite{zhang2023adding}, which shows great control ability given 2D conditions, we adopt a similar principle to train an additional control branch.
To predict the noise $\epsilon_\theta(x_t,t,c)$ in Eq.~\eqref{eq:train_obj}, we encode the corrupted ground-truth shape $x_t$ by a main branch and the partial shape $c$ by a control branch, where two branches have the same network structure without parameter sharing. 
Owing to the compact shape representation in 3D volume space, both complete and incomplete shapes are encoded into multi-resolution feature volumes with preserved spatial structures. Then, we hierarchically aggregate their features at each network level, as the sizes of two feature volumes are always aligned. 
In this work, we simply add up the two feature volumes to allow for a spatially-consistent feature aggregation, \ie, only features at the same 3D location are combined. Compared with frequently-used cross-attention technique~\cite{cheng2022sdfusion,chou2022diffusionsdf,rombach2022high}, doing so can significantly reduce computation costs. By multi-scale feature interaction, the network can effectively correlate the difference in partialness between two shapes, both locally and globally, to learn the completion regularities.
The final integrated features are then used to predict the noise volume $\epsilon_\theta$.

\label{sec:method}
\begin{figure}
	\begin{center}
\includegraphics[width=1\columnwidth]{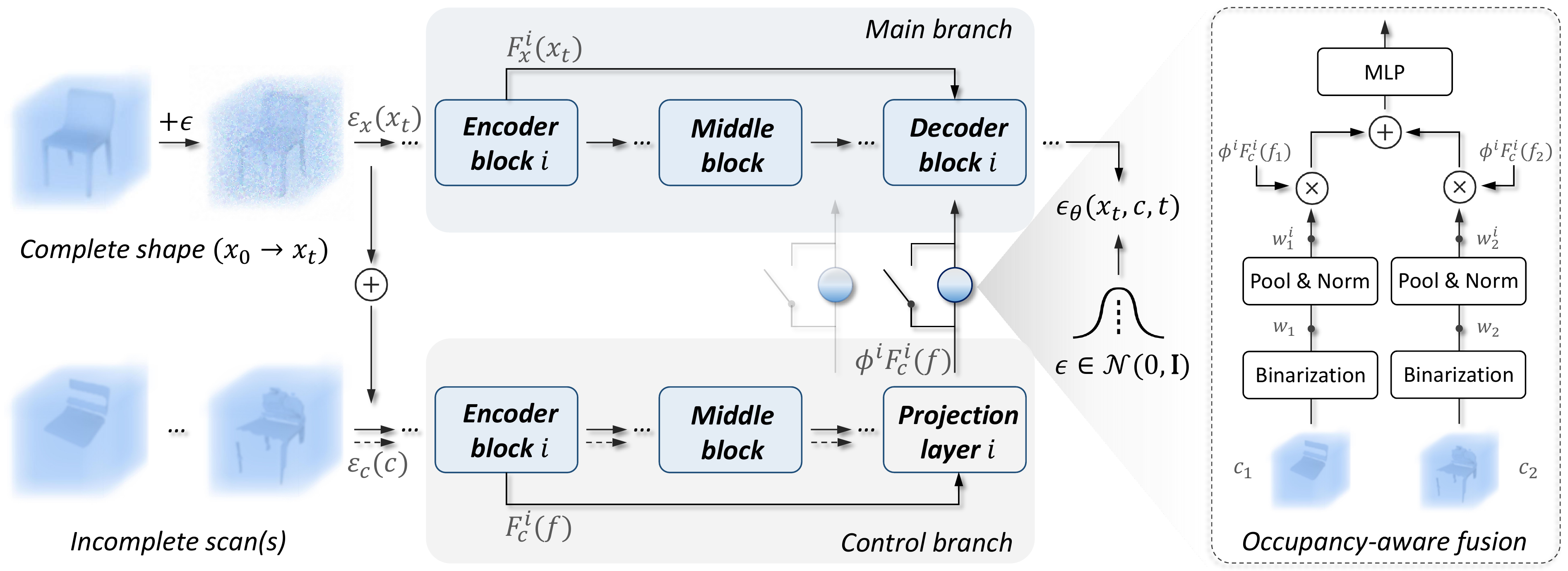}
	\end{center}
	\caption{Given a corrupted complete shape $x_t$ (diffused from $x_0$) and an incomplete scan $c$, we first process them into $\varepsilon_x(x_t)$ and $\varepsilon_c(c)$ to align the distributions. We employ a main branch to forward $\varepsilon_x(x_t)$, and a control branch to propagate their fused features $f$ into deep layers. Multi-level features of $f$ are aggregated into the main branch for hierarchical control in predicting the diffusion noise. To support multiple partial scans as condition, \eg, two scans $\{c_1,c_2\}$, we switch on occupancy-aware fusion (Sec.~\ref{subsec:occ_aware}). This strategy utilizes the occupancy masks to enable a weighted feature fusion for $c_1$ and $c_2$ by considering their geometry reliability before feeding them into the main branch.}
\label{fig:network}
\vspace*{-2mm}
\end{figure}

\textbf{Network architecture.} Fig.~\ref{fig:network} provides an overview of our network architecture, which has a main branch and a control branch to respectively handle complete and incomplete shapes.
The main branch (upper) is a 3D U-Net modified from a 2D version~\cite{nichol2021improved}. It takes as input an corrupted complete shape $x_t$, including (i) a pre-processing block $\varepsilon_x(\cdot)$ of two convolution layers to project $x_t$ into a high-dimension space, (ii) $N$ subsequent encoder blocks, denoted as $\{F_x^i(\cdot)\}_{i=1}^N$,
to progressively encode and downsample the corrupted shape $x_t$ into a collection of multi-scale features, (iii) a middle block $M_x(\cdot)$ with a self-attention layer to incorporate non-local information into the encoded feature volume, and (iv) $N$ decoder blocks $\{D_x^i(\cdot)\}_{i=1}^N$ that sequentially upsample features through an inversion convolution to produce a feature volume of the same size as the input $x_t$. 
The term ``e(d)ncoder/middle block'' denotes a group of neural layers commonly adopted as a unit in neural networks,~\eg, a ``resnet+downsample'' block. 

We use a control branch (bottom) to extract the features of incomplete scan(s).
Likewise, we employ a pre-processing block, $N$ encoder blocks, and a middle block to extract multi-scale features from the conditional input, denoted as $\varepsilon_c(\cdot)$, $\{F_c^i(\cdot)\}_{i=1}^N$, and $M_c$, respectively. They mirror the architecture of the corresponding blocks in the upper branch,
yet operating with non-shared parameters, considering the different input data regularities (\eg, complete and sparse). 
Unlike the upper branch, there are no decoders for computation efficiency.
Instead, we append a projection layer after each encoder/middle block $F_c^i/M_c$ to forward multi-scale features and feed them into the decoder blocks of the main branch.
As diffusion models are time-conditioned, we also convert the time step $t$ into an embedding via two MLPs and add it to the volume features in each network block (the blue box in Fig.~\ref{fig:network}).

\textbf{Hierarchical feature aggregation.} To integrate the features of the complete and incomplete shapes for conditional control, we fuse them across multiple network levels. 
In this process, we first consider a \textit{single} incomplete shape (denoted as $c$) as the condition. Given the significant disparity between $c$ and $x_t$ (incomplete v.s. complete, TSDF v.s. TUDF), we employ the pre-processing layers ($\varepsilon_x(\cdot)$ and $\varepsilon_c(\cdot))$ to first align their distributions before fusing them into feature volume $f$, \ie, $f=\varepsilon_x(x_t)+\varepsilon_c(c)$.
Then, $f$ is passed through the control branch to better propagate the useful information to the deeper layers.
In turn, we use multi-scale condition features from the control branch to guide each level of decoder blocks $\{D_x^i(\cdot)\}_{i=1}^N$ in the main branch.
The features that enter the $i$-th decoder block are denoted as 
\begin{equation}
    d^i=[D^{i-1}_x(x_t), F^i_x(x_t)+\phi^i(F^i_c(f))],\quad f=\varepsilon_x(x_t) + \varepsilon_c(c)
    \label{eq:fuse}
\end{equation}
where $[\cdot,\cdot]$ is the concatenation operation and $\phi^i$ is one $1\times1$ convolution layer for feature projection. For simplicity, we use $F^i_x(x_0)$ to denote the features of the $x_0$ after the $i$-th encoder block $F^i_x(\cdot)$, as $x_0$ is not directly processed by $F^i_x(\cdot)$; the same notation is used for $F^i_c(f)$ and $D^{i-1}_x(x_t)$. 
The feature after the final decoder block is the network output $\epsilon_\theta(x_t,t,c)$.

By such a design, we can hierarchically incorporate conditional features, leveraging both their local and broader contexts to optimize the network outputs. Interestingly, we observe that adjusting the network level for feature aggregation can alter a trade-off between completion accuracy and diversity. Specifically, 
if we only aggregate features at the low-resolution network layers, we might miss finer geometry details present in the higher-resolution layers. This could reduce completion accuracy, as the model has less contextual information to work with. In contrast, if aggregating features at all levels, the effective control might make completion results closely resemble the ground truths, yet lowering the diversity. Further ablation study is provided in Sec.~\ref{subsec:ablation}.

\subsection{Occupancy-aware Fusion}
\label{subsec:occ_aware}
Our framework can also take multiple incomplete scans as inputs. This option not only provides richer information to constrain the completed shape geometry but also enhances the approach's practicality, particularly in scenarios that a single-pass scan may not fully capture the entire object. Our critical design is to effectively fuse multiple partial shape features. As averaging the original TSDF volumes is sensitive to registration noise, we first register multiple partial shapes (\eg, using Fast-Robust-ICP~\cite{zhang2021fast}) and propose an occupancy-aware approach to fuse them in the feature space.

Given a set of incomplete shapes of the same object, denoted as $\{c_1, ..., c_M\}$, we individually feed them into the control branch to produce feature volumes $\{F_c^i(f_1), ..., F_c^i(f_M)\}$ after the $i$-th encoder block. Before feeding them into the decoder $D_x^i$, we refer to their occupancy masks to perform a weighted feature average. 
Concretely, we first compute the original occupancy mask for each partial shape based on the TSDF values, \ie, $w_j=(abs(c_j)<\tau)\in\mathbf{B}^{S\times S\times S}$ for the $j$-th partial shape.
$\tau$ is a pre-defined threshold that assigns the volumes near the object surface as occupied and the rest as unoccupied; $\mathbf{B}$ is a binary tensor.
Then, we perform a pooling operation to resize $w_j$ into $w_j^i$ to align the resolution with feature volume $F_c^i(f_j)$ and normalize $w_j^i$ by $w_j^i=w_j^i/(w_0^i+..+w_M^i)$. For the voxels with zero values across all occupancy masks, we uniformly assign them a 1e-2 value to avoid the division-by-zero issue. We rewrite Eq.~\eqref{eq:fuse}, which is for single partial shape condition, as
\begin{equation}
    d^i=[D^{i-1}_x(x_t), F^i_x(x_t)+\psi(\sum_{j}w_j^i\phi^i(F^i_c(f_j)))],\quad f_j=\varepsilon_x(x_t) + \varepsilon_c(c_j)
    \label{eq:multi_input}
\end{equation}
where $\psi$ is an MLP layer to refine the fused condition features, aiming to mitigate the discrepancies among different partial shape features, as validated in Sec.~\ref{subsec:multi_input}.
The right part of Fig.~\ref{fig:network} illustrates the above process with two incomplete scans as the inputs.

\subsection{Training and Inference}

We first train the network with a single incomplete shape as the conditional input. In this phase, all the network parameters, except the MLP layer $\psi$ for occupancy-aware fusion (in Eq.~\eqref{eq:multi_input}), are trained with the objective in Eq.~\eqref{eq:train_obj}. When the network converges, we lock the optimized network parameters and efficiently finetune the MLP layer $\psi$ with multiple incomplete shapes as input.

At the inference stage, we randomize a 3D noise volume as $x_T$ from the standard Gaussian distribution. The trained completion networks are then employed for $T$ iterations to produce $x_0$ from $x_T$ conditioned on partial shape(s) $c$. The occupancy-aware fusion is activated only for multi-condition completion.
To accelerate the inference process, we adopt a technique from~\cite{song2020denoising} to sub-sample a set of time steps from [1,...,$T$/10] during inference. After obtaining the generated shape volume $x_0$, we extract an explicit 3D mesh using the marching cube algorithm~\cite{lorensen1987marching}.

\section{Experiment}
\label{sec:exp}

\subsection{Experimental Setup}
\label{subsec:setup}

\textbf{Benchmarks.} We evaluate on two large-scale shape completion benchmarks: 3D-EPN~\cite{dai2017shape} and PatchComplete~\cite{rao2022patchcomplete}. 3D-EPN comprises 25,590 object instances of eight classes in ShapeNet~\cite{chang2015shapenet}. 
For each instance, six partial scans of varying completeness are created in the $32^3$ TSDF volumes by virtual scanning; the ground-truth counterpart, represented by $32^3$ TUDF, is obtained by a distance field transform on a 3D scanline method~\cite{amanatides1987fast}. 
While using a similar data generation pipeline, PatchComplete emphasizes completing objects of unseen categories. It includes both the synthetic data from ShapeNet~\cite{chang2015shapenet} and the challenging real data from ScanNet~\cite{dai2017scannet}.
For a fair comparison, we follow their data splits and evaluation metrics, \ie, mean $l_1$ error on the TUDF predictions across all voxels on 3D-EPN, and $l_1$ Chamfer Distance (CD) and Intersection over Union (IoU) between the predicted and ground-truth shapes on PatchComplete.
As these metrics only measure the completion accuracy, we introduce other metrics in Sec.~\ref{subsec:multi_modal} to compare multimodal completion characteristics.

\textbf{Implementation details.} We first train our network using a single partial scan as input by 200k iterations on four RTX3090 GPUs, taking around two days. If multiple conditions are needed, we finetune project layers $\psi$ for additional 50k iterations. Adam optimizer~\cite{kingma2014adam} is employed with a learning rate of 1$e^{-4}$ and the batch size is 32. 
On the 3D-EPN benchmark, we train a specific model for completing shapes of each known category; while on PatchComplete, we merge all object categories to optimize one model to promote general completion learning. Due to the unknown class IDs at test time, no classifier-guided~\cite{dhariwal2021diffusion} or classifier-free~\cite{ho2022classifier} sampling techniques are used in our diffusion model. The truncation distance in TSDF/TUDFs is set as 3 voxel units.
More details about network architecture and experiments are available in the supplementary file. 
Unless otherwise specified, we report the results on \textit{single} partial shape completion.

\begin{table}[t!]
\caption{Quantitative shape completion results on objects of known categories~\cite{dai2017shape}.}
    \centering
    \renewcommand\tabcolsep{8pt}
     \resizebox{\textwidth}{!}{
     \begin{tabular}{l 
 | c | c c c c c c c c }
\toprule 
$l_1$-err.~($\downarrow$) & Avg.~($\downarrow$) & Chair & Table & Sofa & Lamp & Plane & Car & Dresser & Boat\\  
\specialrule{0em}{2pt}{0pt}
\hline
\specialrule{0em}{2pt}{0pt}
3D-EPN~\cite{dai2017shape} & 0.374 & 0.418 & 0.377 & 0.392 & 0.388 & 0.421 & 0.259 & 0.381 & 0.356 \\
SDF-StyleGAN~\cite{zheng2022sdf} & 0.278 & 0.321 & 0.256 & 0.289 & 0.280 & 0.295 & 0.224 & 0.273 & 0.282\\ 
RePaint-3D~\cite{lugmayr2022repaint} & 0.266 & 0.289 & 0.264 & 0.266 & 0.268 & 0.302 & 0.214 & 0.285 & 0.243 \\
AutoSDF~\cite{mittal2022autosdf} & 0.217 & 0.201 & 0.258 & 0.226 & 0.275 & 0.184 & 0.187 & 0.248 & 0.157 \\ 
PatchComplete~\cite{rao2022patchcomplete} & 0.088 & 0.134 & 0.095 & 0.084 & 0.087 & 0.061 & 0.053 & 0.134 & 0.058 \\
\specialrule{0em}{2pt}{0pt}
\hline
\specialrule{0em}{2pt}{0pt}
\ourMethod~\small{(Ours)}  & \textbf{0.053} & \textbf{0.070} & \textbf{0.073} & \textbf{0.061} & \textbf{0.059} & \textbf{0.015} & \textbf{0.025} & \textbf{0.086} & \textbf{0.031} \\
\bottomrule
\end{tabular}
}
\label{tab:epn_bench}
\vspace*{-4mm}
\end{table}

\begin{table}[t!]
    \caption{Shape completion results on synthetic objects~\cite{chang2015shapenet} of unseen categories. $\cdot/\cdot$ means CD/IoU.}
    \centering
    \renewcommand\tabcolsep{4.0pt}
     \resizebox{\textwidth}{!}{
     \begin{tabular}{l 
 | c c c c c c }
\toprule
CD($\downarrow$)/IoU($\uparrow$) & 3D-EPN~\cite{dai2017shape} & Few-Shot~\cite{wallace2019few}& IF-Nets~\cite{chibane2020implicit} & Auto-SDF~\cite{mittal2022autosdf}& PatchComplete~\cite{rao2022patchcomplete} & Ours \\
\specialrule{0em}{2pt}{0pt}
\hline
\specialrule{0em}{2pt}{0pt}

Bag & 5.01 / 73.8 & 8.00 / 56.1 & 4.77 / 69.8 & 5.81 / 56.3 & 3.94 / 77.6 & \textbf{3.86} / \textbf{78.3} \\
Lamp & 8.07 / 47.2 & 15.1 / 25.4 & 5.70 / 50.8 & 6.57 / 39.1 & \textbf{4.68} / 56.4 & 4.80 / \textbf{57.9} \\
Bathtub & 4.21 / 57.9 & 7.05 / 45.7 & 4.72 / 55.0 & 5.17 / 41.0 & 3.78 / 66.3 & \textbf{3.52} / \textbf{68.9} \\
Bed & 5.84 / 58.4 & 10.0 / 39.6 & 5.34 / 60.7 & 6.01 / 44.6 & 4.49 / 66.8 & \textbf{4.16} / \textbf{67.1} \\
Basket & 7.90 / 54.0 & 8.72 / 40.6 & \textbf{4.44} / 50.2 & 6.70 / 39.8 & 5.15 / 61.0 & 4.94 / \textbf{65.5} \\
Printer & 5.15 / 73.6 & 9.26 / 56.7 & 5.83 / 70.5 & 7.52 / 49.9 & 4.63 / \textbf{77.6} & \textbf{4.40} / 76.8 \\
Laptop & 3.90 / 62.0 & 10.4 / 31.3 & 6.47 / 58.3 & 4.81 / 51.1 & 3.77 / 63.8 & \textbf{3.52} / \textbf{67.4} \\
Bench & 4.54 / 48.3 & 8.11 / 27.2 & 5.03 / 49.7 & 4.31 / 39.5 & 3.70 / 53.9 & \textbf{3.56} / \textbf{58.2} \\
\specialrule{0em}{2pt}{0pt}
\hline
\specialrule{0em}{2pt}{0pt}
Avg. & 5.58 / 59.4 & 9.58 / 40.3 & 5.29 / 58.1 & 5.86 / 45.2 & 4.27 / 65.4 & \textbf{4.10} / \textbf{67.5} \\
\bottomrule
\end{tabular}
}
\label{tab:patch_synthetic}
\vspace*{-4mm}
\end{table}

\begin{table}[t!]
    \caption{Shape completion results on real-world objects~\cite{dai2017scannet} of unseen categories. $\cdot/\cdot$ means CD/IoU.}
    \centering
    \renewcommand\tabcolsep{3.0pt}
     \resizebox{\textwidth}{!}{
     \begin{tabular}{l 
 | c c c c c c }
\toprule 
CD($\downarrow$)/IoU($\uparrow$) & 3D-EPN~\cite{dai2017shape} & Few-Shot~\cite{wallace2019few}& IF-Nets~\cite{chibane2020implicit} & Auto-SDF~\cite{mittal2022autosdf}& PatchComplete~\cite{rao2022patchcomplete} & Ours \\
\specialrule{0em}{2pt}{0pt}
\hline
\specialrule{0em}{2pt}{0pt}

Bag & 8.83 / 53.7 & 9.10 / 44.9 & 8.96 / 44.2 & 9.30 /  48.7 & 8.23 / \textbf{58.3} & \textbf{7.05} / 48.5 \\
Lamp & 14.3 / 20.7 & 11.9 / 19.6 & 10.2 / 24.9 & 11.2 / 24.4 & 9.42 / 28.4 & \textbf{6.84} / \textbf{30.5} \\
Bathtub & 7.56 / 41.0 & 7.77 / 38.2 & 7.19 / 39.5 & 7.84 / 36.6 & \textbf{6.77} / 48.0 & 8.22 / \textbf{48.5} \\
Bed & 7.76 / 47.8 & 9.07 / 34.9 & 8.24 / 44.9 & 7.91 / 38.0 & 7.24 / \textbf{48.4} & \textbf{7.20} / 46.6 \\
Basket & 7.74 / 36.5 & 8.02 / 34.3 & 6.74 / 42.7 & 7.54 / 36.1 & \textbf{6.60} / 45.5 & 7.42 / \textbf{59.2} \\
Printer & 8.36 / 63.0 & 8.30 / 62.2 & 8.28 / 60.7 & 9.66 / 49.9 & 6.84 / 70.5 & \textbf{6.36} / \textbf{74.5} \\
\specialrule{0em}{2pt}{0pt}
\hline
\specialrule{0em}{2pt}{0pt}
Avg. & 9.09 / 44.0 & 9.02 / 38.6 & 8.26 / 42.6 & 8.90 / 38.9 & 7.52 / 49.5 & \textbf{7.18} / \textbf{51.3} \\

\bottomrule
\end{tabular}
}
\label{tab:patch_realworld}
\vspace*{-4mm}
\end{table}

\subsection{Main Results}
\label{subsec:main_results}

\begin{figure}
	\begin{center}	\includegraphics[width=0.9\columnwidth]{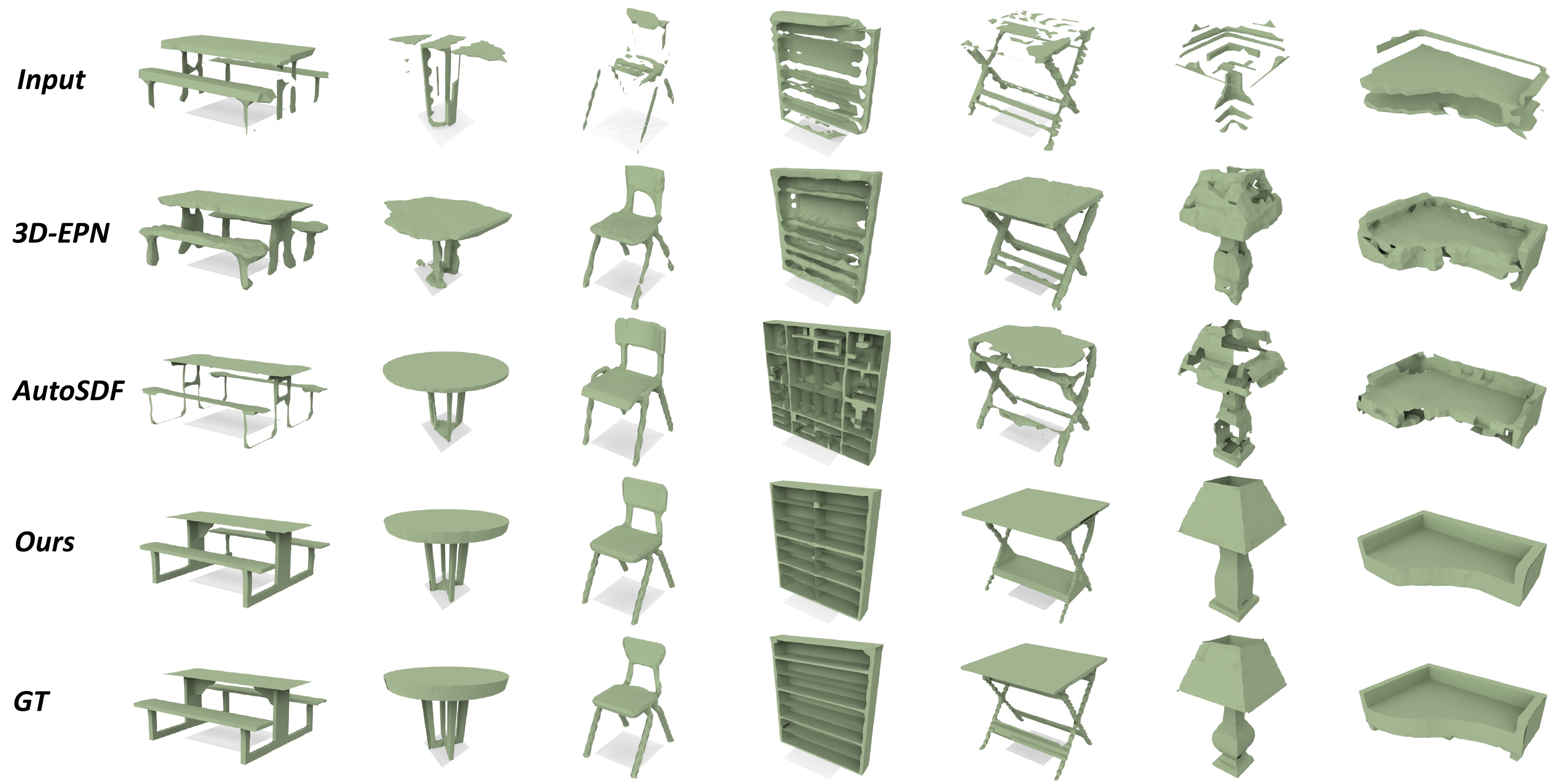}
	\end{center}
	\vspace*{-2mm}
 \caption{Shape Completion on various known object classes. We achieve the best completion quality.}
	\label{fig:vis_main}
\end{figure}

\textbf{Completion on known object categories.} On the 3D-EPN benchmark, we compare \ourMethod~against SOTA deterministic~\cite{dai2017shape,rao2022patchcomplete,zheng2022sdf} and probabilistic~\cite{mittal2022autosdf,lugmayr2022repaint} methods in terms of completion accuracy (\ie, $l_1$ errors).
For probabilistic methods, we use the average results from five inferences, each with random initialization, to account for multimodal outcomes. As shown in Table~\ref{tab:epn_bench} and Fig.~\ref{fig:vis_main}, \ourMethod~improves over state of the arts by 40\% on $l_1$ error (0.053 v.s. 0.088), as well as producing more realistic and high-fidelity shapes.
Unlike 3D-EPN~\cite{dai2017shape} and PatchComplete~\cite{rao2022patchcomplete} that learn a one-step map function for shape completion, we iteratively refine the generated shape, thus significantly mitigating the surface artifacts; see comparisons of 3D-EPN and Ours in Fig.~\ref{fig:vis_main}. 
Compared to GAN-based SDF-StyleGAN~\cite{zheng2022sdf} and AutoEncoding-based AutoSDF~\cite{mittal2022autosdf}, our diffusion model offers superior mode coverage and sampling quality. 
RePaint-3D is adapted from Repaint~\cite{lugmayr2022repaint}, a 2D diffusion-based inpainting method that only involves partial shape conditions during the inference process. In contrast, our~\ourMethod~explicitly matches each partial shape with a complete counterpart at the training stage, thereby improving the output consistency with the ground truths. 

\begin{figure}
	\begin{center}
\includegraphics[width=0.9\columnwidth]{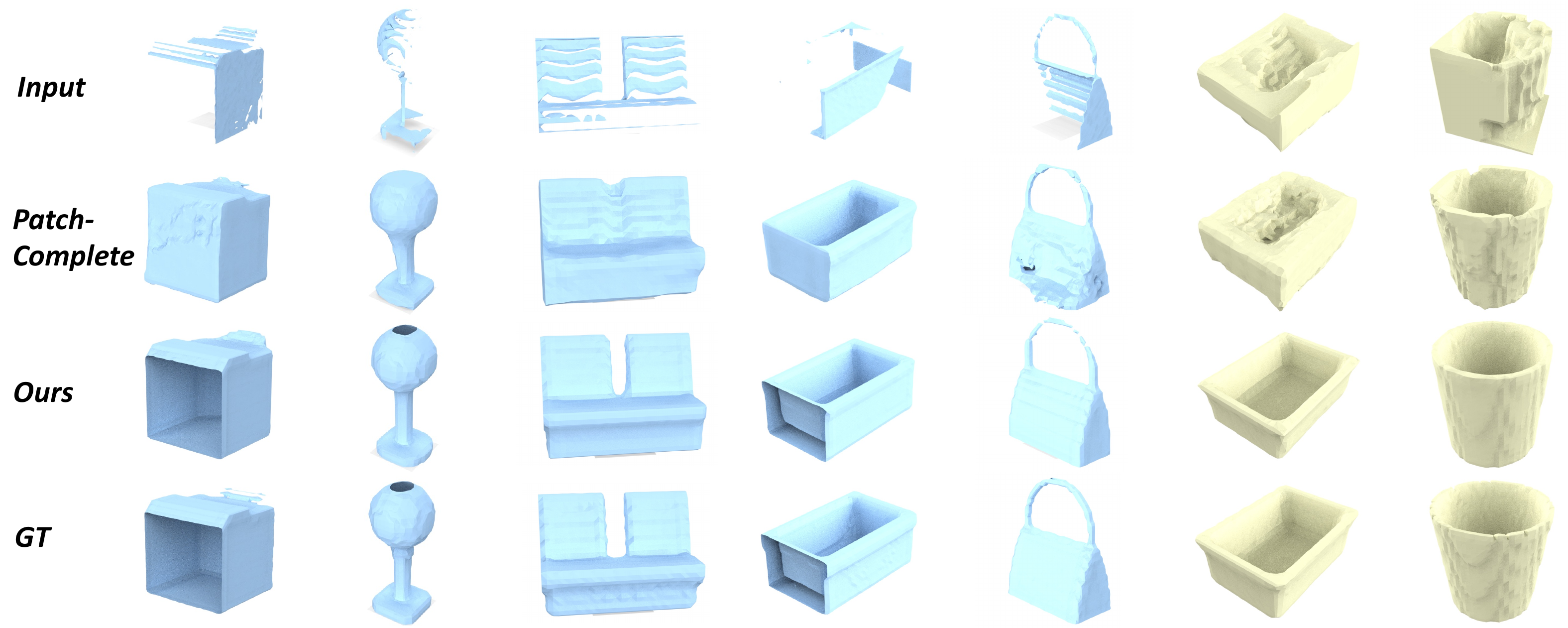}
	\end{center}
	\vspace*{-2mm}
 \caption{Shape completion on synthetic (blue) and real (yellow) objects of \textit{entirely unseen} classes. Our method produces the completed shapes in superior quality given both synthetic and real data.}
	\label{fig:unseen}
\end{figure}

\textbf{Completion on unseen object categories.}
In two datasets of the PatchComplete benchmark, we compare the generalizability of \ourMethod~against the state of the arts, including approaches particularly designed for unseen-class completion~\cite{wallace2019few,rao2022patchcomplete}.
As summarized in Table~\ref{tab:patch_synthetic}, our method exhibits the best completion quality on average for eight unseen object categories in the synthetic ShapeNet data, despite lacking zero-shot designs.
The previous SOTA PatchComplete, mainly leverages the multi-scale structural information to improve the completion robustness. Our method inherently embraces this concept within the diffusion models. With our hierarchical feature aggregation, the network learns multi-scale local completion patterns, which could generalize to various object classes, as their local structures are often shared. Our ablation study in Sec.~\ref{subsec:ablation} further validates this benefit.
Table~\ref{tab:patch_realworld} demonstrates our method's superior performance with real-world scans, which are often cluttered and noisy.
As showcased in Fig.~\ref{fig:unseen}, the 3D shapes produced by \ourMethod~stand out for their impressive global coherence and local details.

\begin{figure}[t!]
    \begin{minipage}{.48\textwidth}
   \centering  \includegraphics[width=1\textwidth]{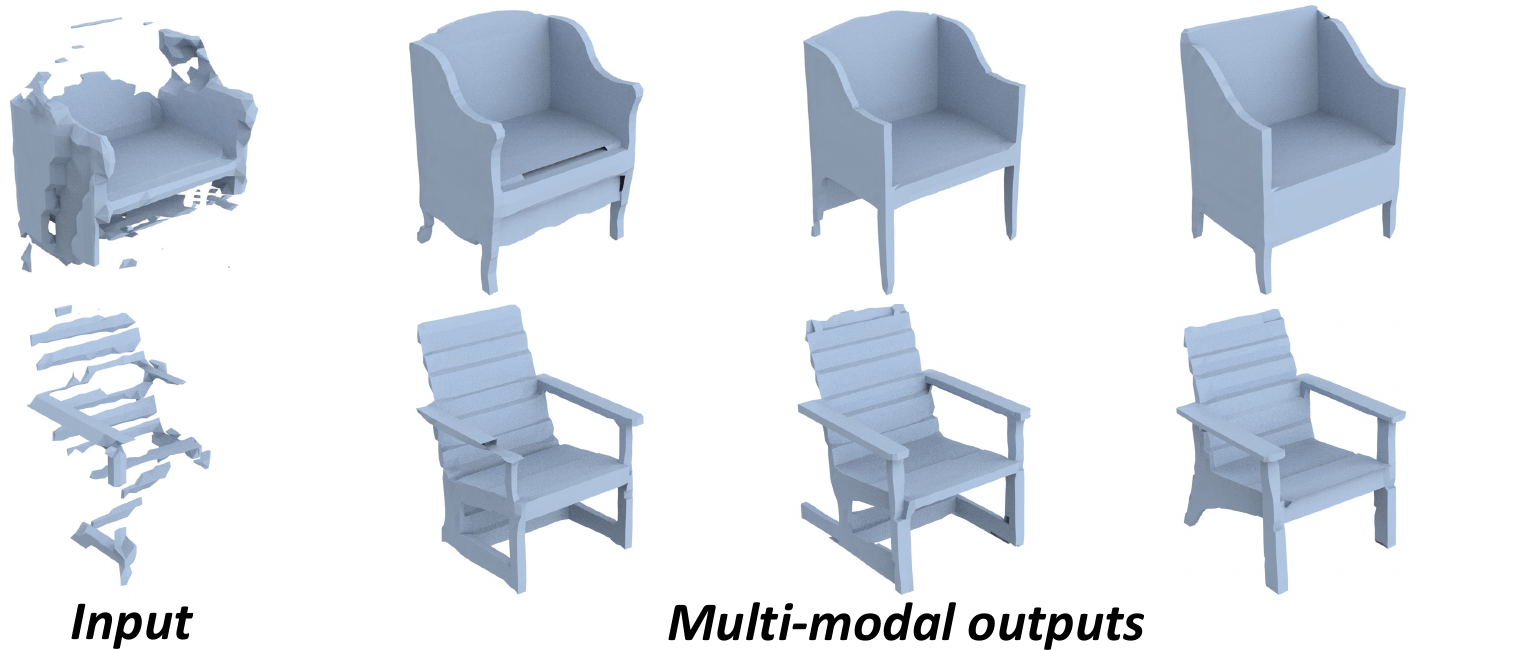}
        \vspace*{-2mm}
        \caption{Our method produces multimodal plausible results given the same partial shape.}
        \vspace*{-2mm}
        \label{fig:multi_modal}
    \end{minipage}
    \hfill
    \hfill
    \begin{minipage}{.48\textwidth}
        \centering
\includegraphics[width=1\textwidth]{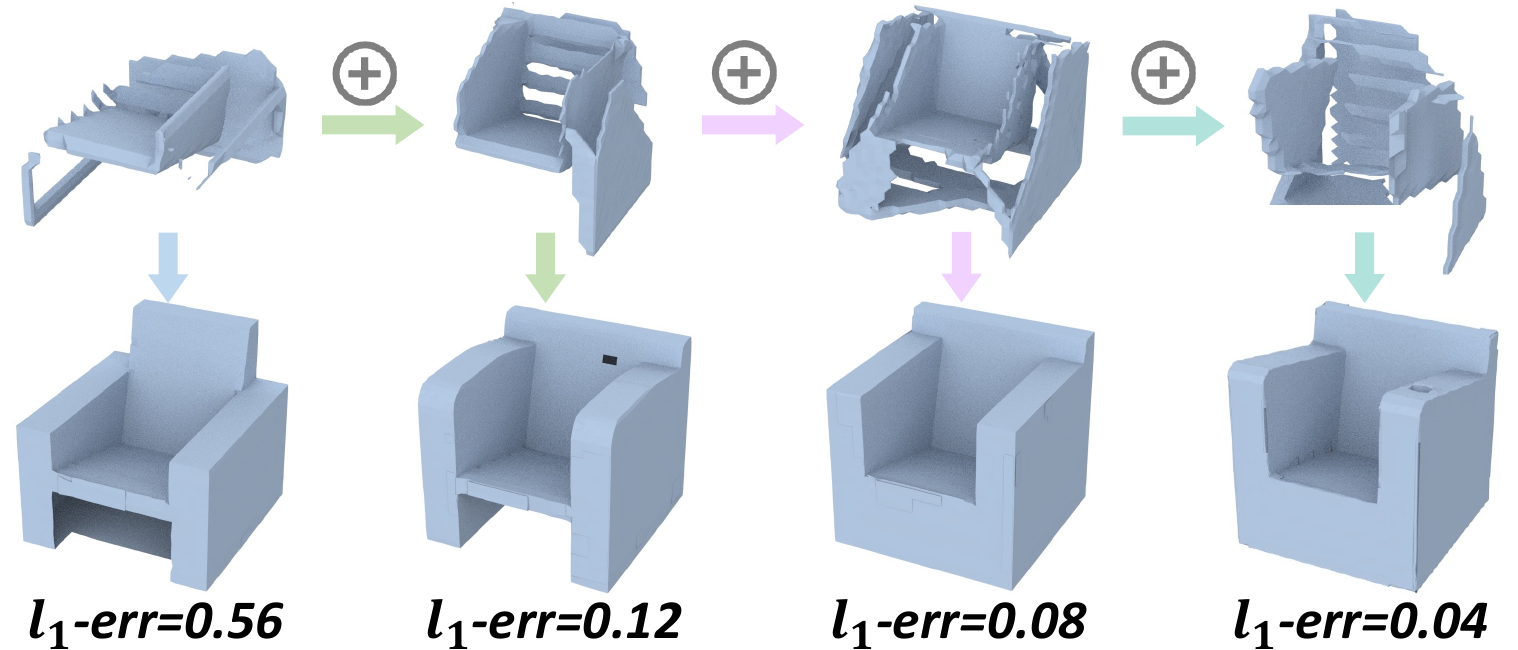} 
        \vspace*{-2mm}
        \caption{Our method incorporates multiple partial shapes to refine completion results ($l_1$-err.~$\downarrow$).}
        \vspace*{-2mm}
        \label{fig:multi_input}
    \end{minipage}
\end{figure}

\subsection{Multimodal Completion Characteristics}
\label{subsec:multi_modal}

\textbf{Quantitative evaluations.} 
Probabilistic methods have the intriguing ability to produce multiple plausible completions on the same partial shape, known as multimodal completion.
For the 3D-EPN chair class, we generate ten results per partial shape with randomized initial noise $x_T$, and employ metrics from prior works~\cite{wu2020multimodal,mittal2022autosdf}, \ie, (i) MMD measures the completion accuracy against the ground truths, (ii) TMD for completion diversity, and (iii) UHD for completion fidelity to a partial input.
Table~\ref{tab:multi_modal} shows that our method attains much better completion accuracy and fidelity, while exhibiting moderate diversity. This aligns with our design choice of leveraging the control mechanism to prioritize completion accuracy over diversity. 
Yet, we can adjust this trade-off to improve shape diversity, as discussed in Sec.~\ref{subsec:ablation}.
Fig.~\ref{fig:multi_modal} presents our multimodal outputs, all showing great realism.

\begin{minipage}{.32\textwidth}
\centering
\makeatletter\def\@captype{table}
\makeatother\caption{Multimodal capacity.}
    \vspace*{-2mm}
    \label{tab:multi_modal}
    \renewcommand\tabcolsep{1pt}
    \resizebox{\textwidth}{!}{
    \begin{tabular}{l| c c c }
    \toprule
    Method & MMD~$\downarrow$ & TMD~$\uparrow$ & UHD~$\downarrow$ \\
    \midrule
    AutoSDF~\cite{mittal2022autosdf} & 0.008 & 0.028 & 0.061 \\
    RePaint-3D~\cite{lugmayr2022repaint} & 0.007 & \textbf{0.029} & 0.053 \\
    Ours & \textbf{0.002} & 0.025 & \textbf{0.032} \\
    \bottomrule
    \end{tabular}
    }
    \end{minipage}
% % % % % % % % % % % % % % % % % % % % 
    \begin{minipage}{.32\textwidth}
    \centering
    \vspace*{2mm}
    \makeatletter\def\@captype{figure}
\makeatother\caption{TMD curve.}
    \vspace*{-3mm}
\includegraphics[width=0.85\textwidth]{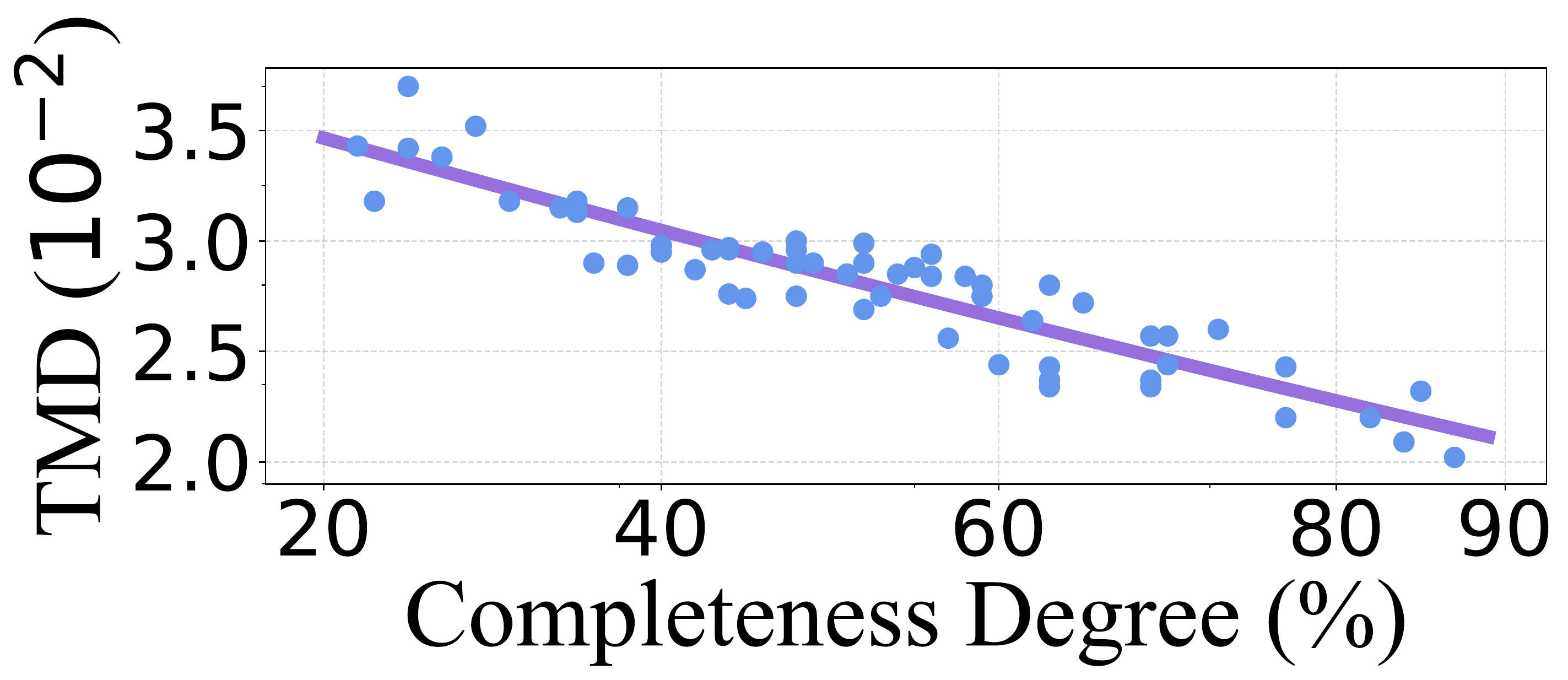}   
        \label{fig:tmd_completeness}
    \end{minipage}
% % % % % % % % % % % % % % % % % % % % 
\begin{minipage}{.32\textwidth}
    \centering
    \vspace*{1mm}
    \makeatletter\def\@captype{table}
\makeatother\caption{Multi-condition results.}
    \vspace*{-2mm}
    \label{tab:multi_condition}
    \renewcommand\tabcolsep{3pt}
    \resizebox{\textwidth}{!}{
    \begin{tabular}{l| c c c }
    \toprule
    Cond. Num. & $l_1$-err.~$\downarrow$ & MMD~$\downarrow$ & TMD~$\uparrow$  \\
    \midrule
    one & 0.07 & 0.002 & \textbf{0.025} \\
    two & 0.05 & 0.001 & 0.021 \\
    three & \textbf{0.04} & \textbf{0.001} & 0.019 \\
    \bottomrule
    \end{tabular}
    }
    \end{minipage}
% % % % % % % % % % % % % % % % % % % % 
\begin{minipage}{.33\textwidth}
    \centering
    \makeatletter\def\@captype{table}
    \vspace*{4mm}
\makeatother\caption{Effects of fusion choices for multiple conditional inputs.}
        \vspace*{-2mm}
    \label{tab:occ_aware}
    \renewcommand\tabcolsep{3pt}
    \resizebox{\textwidth}{!}{
    \begin{tabular}{l| c c c }
    \toprule
    Strategy & $l_1$-err.~$\downarrow$ & CD~$\downarrow$ & IoU~$\uparrow$  \\
    \midrule
    simple average & 0.13 & 4.56 & 63.3 \\
    w/o MLP $\psi$ & 0.23 & 5.88 & 57.4 \\
    occ-aware \small{(Ours)} & \textbf{0.05} & \textbf{3.97} & \textbf{68.3} \\
    \bottomrule
    \end{tabular}
    }
    \end{minipage}
% % % % % % % % % % % % % % % % % % % % 
\begin{minipage}{.33\textwidth}
    \centering
    \makeatletter\def\@captype{table}
        \vspace*{4mm}
\makeatother\caption{Feat. aggregation levels.}
    \vspace*{-2mm}
    \label{tab:fusion_level}
    \renewcommand\tabcolsep{3.5pt}
    \resizebox{\textwidth}{!}{
    \begin{tabular}{ c c c c | c c | c}
    \toprule
    1/8 & 1/4 & 1/2 & 1 & MMD~$\downarrow$ & TMD~$\uparrow$ & CD~$\downarrow$  \\
    \midrule
    \Checkmark &  &  &  & 0.005 & \textbf{0.031} & 4.8 \\
    \Checkmark & \Checkmark &  &  & 0.004 & 0.028 & 4.4\\
    \Checkmark & \Checkmark & \Checkmark &  & 0.003 & 0.026 & 4.2\\
    \Checkmark & \Checkmark  & \Checkmark  & \Checkmark  & \textbf{0.002} & 0.025 & \textbf{4.1} \\
    \bottomrule
    \end{tabular}
    }
    \end{minipage}
\begin{minipage}{.33\textwidth}
    \centering
    \makeatletter\def\@captype{table}
        \vspace*{4mm}
\makeatother\caption{Ablation on different feature aggregation manners.}
    \vspace*{-2mm}
    \label{tab:fusion_opr}
    \renewcommand\tabcolsep{1.7pt}
    \resizebox{\textwidth}{!}{
    \begin{tabular}{l| c c c }
    \toprule
    Operation & $l_1$-err.~$\downarrow$ & MMD~$\downarrow$ & TMD~$\uparrow$  \\
    \midrule
    cross-attn. & 0.12 & 0.005 & \textbf{0.027} \\
    concat. & 0.07 & 0.002 & 0.024 \\
    addition \small{(Ours)} & \textbf{0.07} & \textbf{0.002} & 0.025 \\
    \bottomrule
    \end{tabular}
    }
    \end{minipage}
    
\vspace*{4mm}
\textbf{Effects of input completeness degree.}
To verify the influence of input completeness degree on diversity,
we select ten chair instances from ShapeNet, due to chair's large structural variation.
For each instance, we create six virtual scans of varying completeness, and for each scan, we generate ten diverse completions to compute their MMD.
Fig.~\ref{fig:tmd_completeness} presents a negative correlation between the shape diversity (reflected by MMD) and input completeness degree, which is measured by the occupied voxel's ratio between the partial and complete GT shapes.

\vspace*{-2mm}
\subsection{Multiple Conditional Inputs}
\label{subsec:multi_input}

\textbf{Quantitative evaluations.} 
As indicated in Table~\ref{tab:multi_condition}, \ourMethod~consistently improves completion accuracy (lower $l_1$ error) and reduces diversity (lower TMD) when more conditional inputs are added.
Fig.~\ref{fig:teaser}(b) and~\ref{fig:multi_input} showcase the progression of completion results when we gradually introduce more partial shapes of the same object (denoted by the plus symbol).
Our model incorporates the local structures of all partial inputs and harmonizes them into a coherent final shape with the lower $l_1$ error.

\textbf{Effects of occupancy-aware fusion.} In Table~\ref{tab:occ_aware}, we compare our design with alternatives on the 3D-EPN chair class and PatchComplete benchmark using two conditional inputs.
Averaging features without occupancy masks largely lowers the completion accuracy due to disturbances from non-informative free-space features. Removing the learnable MLP layer $\psi$ hinders the network's adaptation from single to multiple conditions, also worsening the results.
Instead, we adaptively aggregate multi-condition features and refine them to mitigate their discrepancy for reliable completions.

\subsection{Ablation Study}
\label{subsec:ablation}

\textbf{Effects of hierarchical feature aggregation.} Table~\ref{tab:fusion_level} shows the effects of aggregating features of the complete and incomplete shapes at different decoder levels (see Eq.~\eqref{eq:fuse}).
First, increasing feature aggregation layers (from single to hierarchical) consistently boosts the completion accuracy (lower MMD), while decreasing it yields better diversity (higher TMD). 
If connecting features only at the network layer with 1/8 resolution, we achieve the best TMD that surpasses other methods (see Table~\ref{tab:multi_modal}). 
Thus, the accuracy-diversity trade-off can be adjusted by altering the level of feature aggregation. Second, hierarchical feature aggregation facilitates unseen-class completion (lower CD, tested on PatchComplete benchmark). This improvement suggests that leveraging multi-scale structural information from partial inputs enhances the completion robustness.

\textbf{Effects of feature aggregation manner.} 
We ablate on the way to aggregate $F^i_x(x_t)$ and $\phi^i(F^i_c(f)$ in Eq.~\eqref{eq:fuse}.
As Table~\ref{tab:fusion_opr} shows, direct addition achieves the best completion accuracy (the lowest $l_1$-err and MMD), as it only combines features at the same 3D location to precisely correlate their difference for completion. Cross-attention disrupts this spatial consistency and yields less accurate results.
Concatenation has a similar performance with addition, while the latter is more efficient.

\vspace*{-2mm}
\section{Conclusion and Discussion}
\label{sec:conclusion}
We presented \ourMethod, a new diffusion-based approach to enable multimodal, realistic, and high-fidelity 3D shape completion, surpassing 
prior approaches on completion accuracy and quality. 
This success is attributed to two key designs: a hierarchical feature aggregation mechanism for effective conditional control and an occupancy-aware fusion strategy to seamlessly incorporate additional inputs for geometry refinement.
Also, \ourMethod~exhibits robust generalization to unseen object classes for both synthetic and real data, and allows an adjustable balance between completion diversity and accuracy to suit specific needs.
These features position \ourMethod~as a powerful tool in various applications in 3D perception and content creation.
Yet, as with most diffusion models, \ourMethod~requires additional computation due to its multi-step inference. A comprehensive discussion on the limitation and broader impact is provided in the supplementary file.

%%%%%%%%%%%%%%%%%%%%%%%%%%%%%%%%%%%%%%%%%%%%%%%%%%%%%%%%%%%%

\bibliography{egbib}

\begin{thebibliography}{10}

\bibitem{reco_star}
Michael Zollhöfer, Patrick Stotko, Andreas Görlitz, Christian Theobalt,
  Matthias Nießner, Reinhard Klein, and Andreas Kolb.
\newblock State of the art on 3d reconstruction with rgb‐d cameras.
\newblock {\em Computer Graphics Forum}, 37:625--652, 05 2018.

\bibitem{dai2017bundlefusion}
Angela Dai, Matthias Nie{\ss}ner, Michael Zollh{\"{o}}fer, Shahram Izadi, and
  Christian Theobalt.
\newblock Bundlefusion: Real-time globally consistent 3d reconstruction using
  on-the-fly surface reintegration.
\newblock {\em {ACM} Trans. Graph.}, 36(3):24:1--24:18, 2017.

\bibitem{azinovic2022neural}
Dejan Azinovi{\'c}, Ricardo Martin-Brualla, Dan~B Goldman, Matthias
  Nie{\ss}ner, and Justus Thies.
\newblock Neural rgb-d surface reconstruction.
\newblock In {\em Proceedings of the IEEE/CVF Conference on Computer Vision and
  Pattern Recognition}, pages 6290--6301, 2022.

\bibitem{park2019deepsdf}
Jeong~Joon Park, Peter Florence, Julian Straub, Richard Newcombe, and Steven
  Lovegrove.
\newblock Deepsdf: Learning continuous signed distance functions for shape
  representation.
\newblock In {\em Proceedings of the IEEE/CVF conference on computer vision and
  pattern recognition}, pages 165--174, 2019.

\bibitem{yariv2020multiview}
Lior Yariv, Yoni Kasten, Dror Moran, Meirav Galun, Matan Atzmon, Basri Ronen,
  and Yaron Lipman.
\newblock Multiview neural surface reconstruction by disentangling geometry and
  appearance.
\newblock {\em Advances in Neural Information Processing Systems},
  33:2492--2502, 2020.

\bibitem{mittal2022autosdf}
Paritosh Mittal, Yen-Chi Cheng, Maneesh Singh, and Shubham Tulsiani.
\newblock Autosdf: Shape priors for 3d completion, reconstruction and
  generation.
\newblock In {\em Proceedings of the IEEE/CVF Conference on Computer Vision and
  Pattern Recognition}, pages 306--315, 2022.

\bibitem{wu2020multimodal}
Rundi Wu, Xuelin Chen, Yixin Zhuang, and Baoquan Chen.
\newblock Multimodal shape completion via conditional generative adversarial
  networks.
\newblock In {\em Computer Vision--ECCV 2020: 16th European Conference,
  Glasgow, UK, August 23--28, 2020, Proceedings, Part IV 16}, pages 281--296.
  Springer, 2020.

\bibitem{smith2017improved}
Edward~J Smith and David Meger.
\newblock Improved adversarial systems for 3d object generation and
  reconstruction.
\newblock In {\em Conference on Robot Learning}, pages 87--96. PMLR, 2017.

\bibitem{zhang2021unsupervised}
Junzhe Zhang, Xinyi Chen, Zhongang Cai, Liang Pan, Haiyu Zhao, Shuai Yi,
  Chai~Kiat Yeo, Bo~Dai, and Chen~Change Loy.
\newblock Unsupervised 3d shape completion through gan inversion.
\newblock In {\em Proceedings of the IEEE/CVF Conference on Computer Vision and
  Pattern Recognition}, pages 1768--1777, 2021.

\bibitem{muller2022diffrf}
Norman M{\"u}ller, Yawar Siddiqui, Lorenzo Porzi, Samuel~Rota Bul{\`o}, Peter
  Kontschieder, and Matthias Nie{\ss}ner.
\newblock Diffrf: Rendering-guided 3d radiance field diffusion.
\newblock {\em arXiv preprint arXiv:2212.01206}, 2022.

\bibitem{cheng2022sdfusion}
Yen-Chi Cheng, Hsin-Ying Lee, Sergey Tulyakov, Alexander Schwing, and Liangyan
  Gui.
\newblock Sdfusion: Multimodal 3d shape completion, reconstruction, and
  generation.
\newblock {\em arXiv preprint arXiv:2212.04493}, 2022.

\bibitem{zhou20213d}
Linqi Zhou, Yilun Du, and Jiajun Wu.
\newblock 3d shape generation and completion through point-voxel diffusion.
\newblock In {\em Proceedings of the IEEE/CVF International Conference on
  Computer Vision}, pages 5826--5835, 2021.

\bibitem{chou2022diffusionsdf}
Gene Chou, Yuval Bahat, and Felix Heide.
\newblock Diffusionsdf: Conditional generative modeling of signed distance
  functions.
\newblock {\em arXiv preprint arXiv:2211.13757}, 2022.

\bibitem{dai2017shape}
Angela Dai, Charles Ruizhongtai~Qi, and Matthias Nie{\ss}ner.
\newblock Shape completion using 3d-encoder-predictor cnns and shape synthesis.
\newblock In {\em Proceedings of the IEEE conference on computer vision and
  pattern recognition}, pages 5868--5877, 2017.

\bibitem{rao2022patchcomplete}
Yuchen Rao, Yinyu Nie, and Angela Dai.
\newblock Patchcomplete: Learning multi-resolution patch priors for 3d shape
  completion on unseen categories.
\newblock {\em arXiv preprint arXiv:2206.04916}, 2022.

\bibitem{engel2014lsd}
Jakob Engel, Thomas Sch{\"o}ps, and Daniel Cremers.
\newblock Lsd-slam: Large-scale direct monocular slam.
\newblock In {\em Computer Vision--ECCV 2014: 13th European Conference, Zurich,
  Switzerland, September 6-12, 2014, Proceedings, Part II 13}, pages 834--849.
  Springer, 2014.

\bibitem{mur2015orb}
Raul Mur-Artal, Jose Maria~Martinez Montiel, and Juan~D Tardos.
\newblock Orb-slam: a versatile and accurate monocular slam system.
\newblock {\em IEEE transactions on robotics}, 31(5):1147--1163, 2015.

\bibitem{whelan2015elasticfusion}
Thomas Whelan, Stefan Leutenegger, Renato~F. Salas{-}Moreno, Ben Glocker, and
  Andrew~J. Davison.
\newblock Elasticfusion: Dense {SLAM} without {A} pose graph.
\newblock In {\em Robotics: Science and Systems XI, Sapienza University of
  Rome, Rome, Italy, July 13-17, 2015}, 2015.

\bibitem{maier2017intrinsic3d}
R.~Maier, K.~Kim, D.~Cremers, J.~Kautz, and M.~Nie{\ss}ner.
\newblock Intrinsic3d: High-quality {3D} reconstruction by joint appearance and
  geometry optimization with spatially-varying lighting.
\newblock In {\em International Conference on Computer Vision (ICCV)}, Venice,
  Italy, October 2017.

\bibitem{zollhoefer2015shading}
Michael Zollh{\"o}fer, Angela Dai, Matthias Innmann, Chenglei Wu, Marc
  Stamminger, Christian Theobalt, and Matthias Nie{\ss}ner.
\newblock Shading-based refinement on volumetric signed distance functions.
\newblock {\em ACM Transactions on Graphics (TOG)}, 2015.

\bibitem{DBLP:conf/siggraph/CurlessL96}
Brian Curless and Marc Levoy.
\newblock A volumetric method for building complex models from range images.
\newblock In {\em Proceedings of the 23rd Annual Conference on Computer
  Graphics and Interactive Techniques, {SIGGRAPH} 1996, New Orleans, LA, USA,
  August 4-9, 1996}, pages 303--312, 1996.

\bibitem{izadi2011kinectfusion}
Shahram Izadi, David Kim, Otmar Hilliges, David Molyneaux, Richard~A. Newcombe,
  Pushmeet Kohli, Jamie Shotton, Steve Hodges, Dustin Freeman, Andrew~J.
  Davison, and Andrew~W. Fitzgibbon.
\newblock Kinectfusion: real-time 3d reconstruction and interaction using a
  moving depth camera.
\newblock In {\em Proceedings of the 24th Annual {ACM} Symposium on User
  Interface Software and Technology, Santa Barbara, CA, USA, October 16-19,
  2011}, pages 559--568, 2011.

\bibitem{newcombe2011kinectfusion}
Richard~A. Newcombe, Shahram Izadi, Otmar Hilliges, David Molyneaux, David Kim,
  Andrew~J. Davison, Pushmeet Kohli, Jamie Shotton, Steve Hodges, and Andrew~W.
  Fitzgibbon.
\newblock Kinectfusion: Real-time dense surface mapping and tracking.
\newblock In {\em 10th {IEEE} International Symposium on Mixed and Augmented
  Reality, {ISMAR} 2011, Basel, Switzerland, October 26-29, 2011}, pages
  127--136, 2011.

\bibitem{niessner2013real}
Matthias Nie{\ss}ner, Michael Zollh{\"o}fer, Shahram Izadi, and Marc
  Stamminger.
\newblock Real-time 3d reconstruction at scale using voxel hashing.
\newblock {\em ACM Transactions on Graphics (ToG)}, 32(6):1--11, 2013.

\bibitem{weder2020routedfusion}
Silvan Weder, Johannes Schonberger, Marc Pollefeys, and Martin~R Oswald.
\newblock Routedfusion: Learning real-time depth map fusion.
\newblock In {\em Proceedings of the IEEE/CVF Conference on Computer Vision and
  Pattern Recognition}, pages 4887--4897, 2020.

\bibitem{peng2020convolutional}
Songyou Peng, Michael Niemeyer, Lars Mescheder, Marc Pollefeys, and Andreas
  Geiger.
\newblock Convolutional occupancy networks.
\newblock In {\em Computer Vision--ECCV 2020: 16th European Conference,
  Glasgow, UK, August 23--28, 2020, Proceedings, Part III 16}, pages 523--540.
  Springer, 2020.

\bibitem{li2020multi}
Ang Li, Zejian Yuan, Yonggen Ling, Wanchao Chi, Chong Zhang, et~al.
\newblock A multi-scale guided cascade hourglass network for depth completion.
\newblock In {\em Proceedings of the IEEE/CVF Winter Conference on Applications
  of Computer Vision}, pages 32--40, 2020.

\bibitem{dai2021spsg}
Angela Dai, Yawar Siddiqui, Justus Thies, Julien Valentin, and Matthias
  Nie{\ss}ner.
\newblock Spsg: Self-supervised photometric scene generation from rgb-d scans.
\newblock In {\em Proceedings of the IEEE/CVF Conference on Computer Vision and
  Pattern Recognition}, pages 1747--1756, 2021.

\bibitem{sorkine2004least}
Olga Sorkine and Daniel Cohen-Or.
\newblock Least-squares meshes.
\newblock In {\em Proceedings Shape Modeling Applications, 2004.}, pages
  191--199. IEEE, 2004.

\bibitem{nealen2006laplacian}
Andrew Nealen, Takeo Igarashi, Olga Sorkine, and Marc Alexa.
\newblock Laplacian mesh optimization.
\newblock In {\em Proceedings of the 4th international conference on Computer
  graphics and interactive techniques in Australasia and Southeast Asia}, pages
  381--389, 2006.

\bibitem{zhao2007robust}
Wei Zhao, Shuming Gao, and Hongwei Lin.
\newblock A robust hole-filling algorithm for triangular mesh.
\newblock {\em The Visual Computer}, 23:987--997, 2007.

\bibitem{kazhdan2006poisson}
Michael~M. Kazhdan, Matthew Bolitho, and Hugues Hoppe.
\newblock Poisson surface reconstruction.
\newblock In {\em Proceedings of the Fourth Eurographics Symposium on Geometry
  Processing, Cagliari, Sardinia, Italy, June 26-28, 2006}, pages 61--70, 2006.

\bibitem{kazhdan2013screened}
Michael Kazhdan and Hugues Hoppe.
\newblock Screened poisson surface reconstruction.
\newblock {\em ACM Transactions on Graphics (ToG)}, 32(3):1--13, 2013.

\bibitem{thrun2005shape}
Sebastian Thrun and Ben Wegbreit.
\newblock Shape from symmetry.
\newblock In {\em Tenth IEEE International Conference on Computer Vision
  (ICCV'05) Volume 1}, volume~2, pages 1824--1831. IEEE, 2005.

\bibitem{mitra2006partial}
Niloy~J Mitra, Leonidas~J Guibas, and Mark Pauly.
\newblock Partial and approximate symmetry detection for 3d geometry.
\newblock {\em ACM Transactions on Graphics (ToG)}, 25(3):560--568, 2006.

\bibitem{pauly2008discovering}
Mark Pauly, Niloy~J Mitra, Johannes Wallner, Helmut Pottmann, and Leonidas~J
  Guibas.
\newblock Discovering structural regularity in 3d geometry.
\newblock In {\em ACM SIGGRAPH 2008 papers}. 2008.

\bibitem{sipiran2014approximate}
Ivan Sipiran, Robert Gregor, and Tobias Schreck.
\newblock Approximate symmetry detection in partial 3d meshes.
\newblock In {\em Computer Graphics Forum}, 2014.

\bibitem{speciale2016symmetry}
Pablo Speciale, Martin~R Oswald, Andrea Cohen, and Marc Pollefeys.
\newblock A symmetry prior for convex variational 3d reconstruction.
\newblock In {\em Computer Vision--ECCV 2016: 14th European Conference,
  Amsterdam, The Netherlands, October 11-14, 2016, Proceedings, Part VIII 14},
  pages 313--328. Springer, 2016.

\bibitem{sung2015data}
Minhyuk Sung, Vladimir~G Kim, Roland Angst, and Leonidas Guibas.
\newblock Data-driven structural priors for shape completion.
\newblock {\em ACM Transactions on Graphics (TOG)}, 34(6):1--11, 2015.

\bibitem{li2015database}
Yangyan Li, Angela Dai, Leonidas Guibas, and Matthias Nie{\ss}ner.
\newblock Database-assisted object retrieval for real-time 3d reconstruction.
\newblock In {\em Computer graphics forum}, 2015.

\bibitem{nan2012search}
Liangliang Nan, Ke~Xie, and Andrei Sharf.
\newblock A search-classify approach for cluttered indoor scene understanding.
\newblock {\em ACM Transactions on Graphics (TOG)}, 31(6):1--10, 2012.

\bibitem{kim2012acquiring}
Young~Min Kim, Niloy~J Mitra, Dong-Ming Yan, and Leonidas Guibas.
\newblock Acquiring 3d indoor environments with variability and repetition.
\newblock {\em ACM Transactions on Graphics (TOG)}, 31(6):1--11, 2012.

\bibitem{nguyen2016field}
Duc~Thanh Nguyen, Binh-Son Hua, Khoi Tran, Quang-Hieu Pham, and Sai-Kit Yeung.
\newblock A field model for repairing 3d shapes.
\newblock In {\em Proceedings of the IEEE Conference on Computer Vision and
  Pattern Recognition}, pages 5676--5684, 2016.

\bibitem{firman2016structured}
Michael Firman, Oisin Mac~Aodha, Simon Julier, and Gabriel~J Brostow.
\newblock Structured prediction of unobserved voxels from a single depth image.
\newblock In {\em Proceedings of the IEEE Conference on Computer Vision and
  Pattern Recognition}, pages 5431--5440, 2016.

\bibitem{dai2020sg}
Angela Dai, Christian Diller, and Matthias Nie{\ss}ner.
\newblock Sg-nn: Sparse generative neural networks for self-supervised scene
  completion of rgb-d scans.
\newblock In {\em Proceedings of the IEEE/CVF Conference on Computer Vision and
  Pattern Recognition}, pages 849--858, 2020.

\bibitem{yu2021pointr}
Xumin Yu, Yongming Rao, Ziyi Wang, Zuyan Liu, Jiwen Lu, and Jie Zhou.
\newblock Pointr: Diverse point cloud completion with geometry-aware
  transformers.
\newblock In {\em Proceedings of the IEEE/CVF international conference on
  computer vision}, pages 12498--12507, 2021.

\bibitem{han2017high}
Xiaoguang Han, Zhen Li, Haibin Huang, Evangelos Kalogerakis, and Yizhou Yu.
\newblock High-resolution shape completion using deep neural networks for
  global structure and local geometry inference.
\newblock In {\em Proceedings of the IEEE international conference on computer
  vision}, pages 85--93, 2017.

\bibitem{song2017semantic}
Shuran Song, Fisher Yu, Andy Zeng, Angel~X Chang, Manolis Savva, and Thomas
  Funkhouser.
\newblock Semantic scene completion from a single depth image.
\newblock In {\em Proceedings of the IEEE conference on computer vision and
  pattern recognition}, pages 1746--1754, 2017.

\bibitem{chibane2020implicit}
Julian Chibane, Thiemo Alldieck, and Gerard Pons-Moll.
\newblock Implicit functions in feature space for 3d shape reconstruction and
  completion.
\newblock In {\em Proceedings of the IEEE/CVF conference on computer vision and
  pattern recognition}, pages 6970--6981, 2020.

\bibitem{dai2019scan2mesh}
Angela Dai and Matthias Nie{\ss}ner.
\newblock Scan2mesh: From unstructured range scans to 3d meshes.
\newblock In {\em Proceedings of the IEEE/CVF Conference on Computer Vision and
  Pattern Recognition}, pages 5574--5583, 2019.

\bibitem{zheng2022sdf}
X~Zheng, Yang Liu, P~Wang, and Xin Tong.
\newblock Sdf-stylegan: Implicit sdf-based stylegan for 3d shape generation.
\newblock In {\em Computer Graphics Forum}, 2022.

\bibitem{chen2019unpaired}
Xuelin Chen, Baoquan Chen, and Niloy~J Mitra.
\newblock Unpaired point cloud completion on real scans using adversarial
  training.
\newblock {\em arXiv preprint arXiv:1904.00069}, 2019.

\bibitem{achlioptas2018learning}
Panos Achlioptas, Olga Diamanti, Ioannis Mitliagkas, and Leonidas Guibas.
\newblock Learning representations and generative models for 3d point clouds.
\newblock In {\em International conference on machine learning}, pages 40--49.
  PMLR, 2018.

\bibitem{sohl2015deep}
Jascha Sohl-Dickstein, Eric Weiss, Niru Maheswaranathan, and Surya Ganguli.
\newblock Deep unsupervised learning using nonequilibrium thermodynamics.
\newblock In {\em International Conference on Machine Learning}, pages
  2256--2265. PMLR, 2015.

\bibitem{dhariwal2021diffusion}
Prafulla Dhariwal and Alexander Nichol.
\newblock Diffusion models beat gans on image synthesis.
\newblock {\em Advances in Neural Information Processing Systems},
  34:8780--8794, 2021.

\bibitem{rombach2022high}
Robin Rombach, Andreas Blattmann, Dominik Lorenz, Patrick Esser, and Bj{\"o}rn
  Ommer.
\newblock High-resolution image synthesis with latent diffusion models.
\newblock In {\em Proceedings of the IEEE/CVF Conference on Computer Vision and
  Pattern Recognition}, pages 10684--10695, 2022.

\bibitem{song2020score}
Yang Song, Jascha Sohl-Dickstein, Diederik~P Kingma, Abhishek Kumar, Stefano
  Ermon, and Ben Poole.
\newblock Score-based generative modeling through stochastic differential
  equations.
\newblock {\em ICLR}, 2021.

\bibitem{nichol2021improved}
Alexander~Quinn Nichol and Prafulla Dhariwal.
\newblock Improved denoising diffusion probabilistic models.
\newblock In {\em International Conference on Machine Learning}, pages
  8162--8171. PMLR, 2021.

\bibitem{ho2020denoising}
Jonathan Ho, Ajay Jain, and Pieter Abbeel.
\newblock Denoising diffusion probabilistic models.
\newblock {\em Advances in Neural Information Processing Systems},
  33:6840--6851, 2020.

\bibitem{song2020denoising}
Jiaming Song, Chenlin Meng, and Stefano Ermon.
\newblock Denoising diffusion implicit models.
\newblock {\em arXiv preprint arXiv:2010.02502}, 2020.

\bibitem{sinha2021d2c}
Abhishek Sinha, Jiaming Song, Chenlin Meng, and Stefano Ermon.
\newblock D2c: Diffusion-decoding models for few-shot conditional generation.
\newblock {\em Advances in Neural Information Processing Systems},
  34:12533--12548, 2021.

\bibitem{zeng2022lion}
Xiaohui Zeng, Arash Vahdat, Francis Williams, Zan Gojcic, Or~Litany, Sanja
  Fidler, and Karsten Kreis.
\newblock Lion: Latent point diffusion models for 3d shape generation.
\newblock {\em arXiv preprint arXiv:2210.06978}, 2022.

\bibitem{luo2021diffusion}
Shitong Luo and Wei Hu.
\newblock Diffusion probabilistic models for 3d point cloud generation.
\newblock In {\em Proceedings of the IEEE/CVF Conference on Computer Vision and
  Pattern Recognition}, pages 2837--2845, 2021.

\bibitem{nichol2022point}
Alex Nichol, Heewoo Jun, Prafulla Dhariwal, Pamela Mishkin, and Mark Chen.
\newblock Point-e: A system for generating 3d point clouds from complex
  prompts.
\newblock {\em arXiv preprint arXiv:2212.08751}, 2022.

\bibitem{nam20223d}
Gimin Nam, Mariem Khlifi, Andrew Rodriguez, Alberto Tono, Linqi Zhou, and Paul
  Guerrero.
\newblock 3d-ldm: Neural implicit 3d shape generation with latent diffusion
  models.
\newblock {\em arXiv preprint arXiv:2212.00842}, 2022.

\bibitem{zhang20233dshape2vecset}
Biao Zhang, Jiapeng Tang, Matthias Niessner, and Peter Wonka.
\newblock 3dshape2vecset: A 3d shape representation for neural fields and
  generative diffusion models.
\newblock {\em arXiv preprint arXiv:2301.11445}, 2023.

\bibitem{bautista2022gaudi}
Miguel~Angel Bautista, Pengsheng Guo, Samira Abnar, Walter Talbott, Alexander
  Toshev, Zhuoyuan Chen, Laurent Dinh, Shuangfei Zhai, Hanlin Goh, Daniel
  Ulbricht, et~al.
\newblock Gaudi: A neural architect for immersive 3d scene generation.
\newblock {\em Advances in Neural Information Processing Systems},
  35:25102--25116, 2022.

\bibitem{zhang2023adding}
Lvmin Zhang and Maneesh Agrawala.
\newblock Adding conditional control to text-to-image diffusion models.
\newblock {\em arXiv preprint arXiv:2302.05543}, 2023.

\bibitem{zhang2021fast}
Juyong Zhang, Yuxin Yao, and Bailin Deng.
\newblock Fast and robust iterative closest point.
\newblock {\em IEEE Transactions on Pattern Analysis and Machine Intelligence},
  2021.

\bibitem{lorensen1987marching}
William~E Lorensen and Harvey~E Cline.
\newblock Marching cubes: A high resolution 3d surface construction algorithm.
\newblock {\em ACM siggraph computer graphics}, 21(4):163--169, 1987.

\bibitem{chang2015shapenet}
Angel~X Chang, Thomas Funkhouser, Leonidas Guibas, Pat Hanrahan, Qixing Huang,
  Zimo Li, Silvio Savarese, Manolis Savva, Shuran Song, Hao Su, et~al.
\newblock Shapenet: An information-rich 3d model repository.
\newblock {\em arXiv preprint arXiv:1512.03012}, 2015.

\bibitem{amanatides1987fast}
John Amanatides, Andrew Woo, et~al.
\newblock A fast voxel traversal algorithm for ray tracing.
\newblock In {\em Eurographics}, 1987.

\bibitem{dai2017scannet}
Angela Dai, Angel~X Chang, Manolis Savva, Maciej Halber, Thomas Funkhouser, and
  Matthias Nie{\ss}ner.
\newblock Scannet: Richly-annotated 3d reconstructions of indoor scenes.
\newblock In {\em Proceedings of the IEEE conference on computer vision and
  pattern recognition}, pages 5828--5839, 2017.

\bibitem{kingma2014adam}
Diederik~P Kingma and Jimmy Ba.
\newblock Adam: A method for stochastic optimization.
\newblock {\em arXiv preprint arXiv:1412.6980}, 2014.

\bibitem{ho2022classifier}
Jonathan Ho and Tim Salimans.
\newblock Classifier-free diffusion guidance.
\newblock {\em arXiv preprint arXiv:2207.12598}, 2022.

\bibitem{lugmayr2022repaint}
Andreas Lugmayr, Martin Danelljan, Andres Romero, Fisher Yu, Radu Timofte, and
  Luc Van~Gool.
\newblock Repaint: Inpainting using denoising diffusion probabilistic models.
\newblock In {\em Proceedings of the IEEE/CVF Conference on Computer Vision and
  Pattern Recognition}, pages 11461--11471, 2022.

\bibitem{wallace2019few}
Bram Wallace and Bharath Hariharan.
\newblock Few-shot generalization for single-image 3d reconstruction via
  priors.
\newblock In {\em Proceedings of the IEEE/CVF International Conference on
  Computer Vision}, pages 3818--3827, 2019.

\bibitem{zheng2022fast}
Hongkai Zheng, Weili Nie, Arash Vahdat, Kamyar Azizzadenesheli, and Anima
  Anandkumar.
\newblock Fast sampling of diffusion models via operator learning.
\newblock {\em arXiv preprint arXiv:2211.13449}, 2022.

\bibitem{3DSemanticSegmentationWithSubmanifoldSparseConvNet}
Benjamin Graham, Martin Engelcke, and Laurens van~der Maaten.
\newblock 3d semantic segmentation with submanifold sparse convolutional
  networks.
\newblock {\em CVPR}, 2018.

\bibitem{wang2017cnn}
Peng-Shuai Wang, Yang Liu, Yu-Xiao Guo, Chun-Yu Sun, and Xin Tong.
\newblock O-cnn: Octree-based convolutional neural networks for 3d shape
  analysis.
\newblock {\em ACM Transactions On Graphics (TOG)}, 36(4):1--11, 2017.

\end{thebibliography}
\bibliographystyle{unsrt}

\clearpage
\appendix

\hypersetup{
    linkcolor=black,
}

\begin{center}
 \LARGE \bf {\textit{Supplementary Material}}
\end{center}
\etocdepthtag.toc{mtappendix}
\etocsettagdepth{mtchapter}{none}
\etocsettagdepth{mtappendix}{subsection}
\tableofcontents

\hypersetup{
    linkcolor=red,
}

\section{Detailed Network Architecture}
\label{sec:supp_detail}

\begin{figure}[htbp]
	\begin{center}
  \includegraphics[width=1.0\columnwidth]{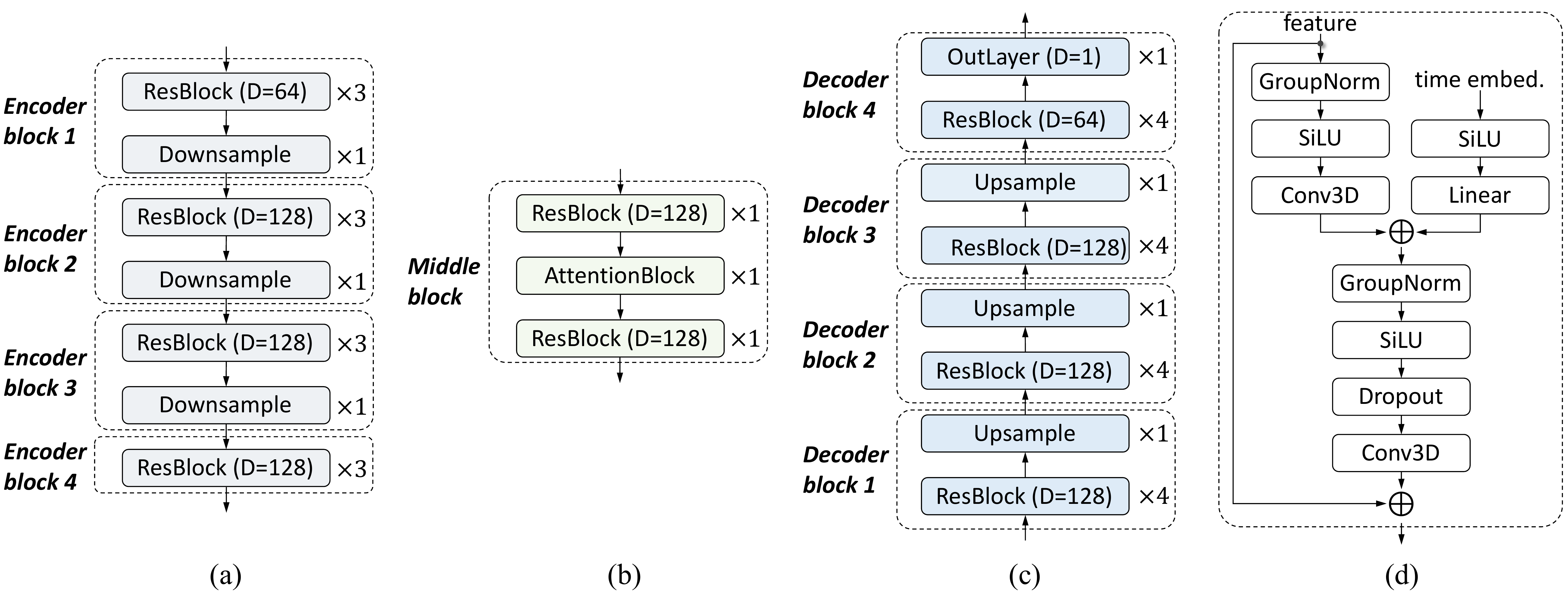}
	\end{center}
	\caption{The detailed architecture of encoder blocks (a), middle blocks (b), and decoder blocks (c) in our network, each is mainly constructed by stacked ResBlocks (d). The `D' denotes the output feature dimension and $\bigoplus$ represents the feature addition operation.}
	\label{fig:supp_network}
\end{figure}

Fig.~\ref{fig:supp_network} shows the detailed architecture of encoder blocks, middle blocks, and decoder blocks of our network (corresponding to those in Fig.~\textcolor{red}{2}). In particular, both the main and control branches consist of four encoder blocks (Fig.~\ref{fig:supp_network}(a)), built from repeated ResBlocks and Downsample layers, where the latter iteratively reduces the feature volume to 1/8th of its original size.
The middle block (Fig.~\ref{fig:supp_network}(b)) comprises two ResBlocks with an intermediate AttentionBlock.
The main branch additionally contains four decoder blocks (Fig.~\ref{fig:supp_network}(c)), which restore the volume shape to its initial size using upsampling.
Fig.~\ref{fig:supp_network}(d) presents the detailed structure of a ResBlock unit. It receives features from the preceding network layer and a time embedding as inputs, fuses their embeddings, and processes them with convolutional operations.
To support network and experiment reproduction, we will make our code available.

\section{Additional Experiments}
\label{sec:supp_exp}

\subsection{Choice of Training Strategy}
\label{subsec:supp_training}

In Table~\ref{tab:supp_training}, we evaluate different training strategies for our model on chair class of the 3D-EPN~\cite{dai2017shape} benchmark.
Both `pretraining-class' and `pretraining-all' follow ControlNet~\cite{zhang2023adding}'s training paradigm.
They involve an initial unconditional generation task for main branch pretraining and then optimize only the control branch with partial shapes with the objective in Eq.~(\textcolor{red}{3}).
Specifically, `pretraining-class' pretrains the main branch on individual object classes, while `pretraining-all' employs all data for one pretraining model, both of which are finetuned on a specific object class.
These pretraining-based strategies bring much more completion errors (higher $l_1$-err.) and increased completion diversity (higher TMD).
This may be due to the model's over-relies on the learned distribution during pretraining, making it less adaptable to concrete completion tasks.
Instead, we train the network entirely from scratch, which is most effective for accurate shape completion (with the lowest $l_1$-err.).

\subsection{Choice of Fusion Space for Multiple Conditions}
\label{subsec:supp_fusion_space}

Table~\ref{tab:supp_fusion} analyzes the impacts of fusion space when incorporating multiple incomplete shapes as conditions, which supplements Table~\textcolor{red}{6} to further validate our fusion choice. The experiments are conducted on the 3D-EPN~\cite{dai2017shape} and Patchcomplete~\cite{rao2022patchcomplete} benchmarks.
The first two rows refer to merging multiple aligned partial shapes into a volumetric TSDF and feeding it to the network as a single condition. However, both 'simple average' and `occupancy-aware' fusion strategies for this yield worse performance across diverse metrics. This is because registration errors between different shapes directly disrupt the input, potentially causing distortions in the final completed model.
In contrast, we move the fusion process to a more abstract level within the hierarchical feature space, which can be more resilient to simple noise at the TSDF level.
Our occupancy-aware fusion strategy further provides more accurate and robust completion results.

\begin{table}[t!]
\begin{minipage}{0.44\textwidth}
\flushleft
\caption{Choice of training strategy. The pretraining-based options increase the completion errors ($l_1$-err.) and diversity (TMD).}
\vspace*{4mm}
\label{tab:supp_training}
    \renewcommand\tabcolsep{6pt}
    \resizebox{\textwidth}{!}{
    \begin{tabular}{l| c c }
    \toprule
    Training Strategy & $l_1$-err.~$\downarrow$ & TMD~$\uparrow$ \\
    \midrule
    pretraining-class & 0.14 & 0.028 \\
    pretraining-all & 0.11 & \textbf{0.030} \\
    scratch~\small{(Ours)} & \textbf{0.07} & 0.025 \\
    \bottomrule
    \end{tabular}
    }
\end{minipage}
\hfill
\begin{minipage}{0.53\textwidth}
\flushright
\caption{Choice of fusion space for multiple partial shapes. Directly fusing them in the original TSDF space significantly impairs the completion quality.}
    \label{tab:supp_fusion}
    \vspace*{1mm}
    \renewcommand\tabcolsep{6pt}
    \resizebox{1.0\textwidth}{!}{
    \begin{tabular}{l| l | c c c}
    \toprule
    Space & Strategy & $l_1$-err.~$\downarrow$ & CD~$\downarrow$ & IoU~$\uparrow$\\
    \midrule
     & simple average & 0.29 & 6.68 & 51.6 \\
    \multirow{-2}{*}{TSDF} & occ-aware & 0.12 & 4.78 & 61.0\\
    \midrule
     & simple average & 0.13 & 4.56 & 63.3 \\
    \multirow{-2}{*}{feature} & occ-aware & \textbf{0.05} & \textbf{3.97} & \textbf{68.3} \\
    \bottomrule
    \end{tabular}
    }
\end{minipage}
\end{table}

\subsection{Choice of Occupancy Threshold}
\label{subsec:supp_thres}

During the occupancy-aware fusion process, the TSDF value threshold $\tau$ determines which volumes are recognized as occupied.
In Table~\ref{tab:supp_threshold}, we evaluate the impact of different thresholds on the completion performance using the 3D-EPN~\cite{dai2017shape} and Patchcomplete~\cite{rao2022patchcomplete} benchmarks.
Selecting extreme threshold values, whether very low (\eg,
1 voxel unit) or high (\eg, 5 voxel units), tends to degrade results, as lower thresholds may omit informative geometries while higher ones could include redundant geometries that do not contribute meaningfully to the object shape, both of which confuse the model.
Conversely, a middle-range value (3 voxel unit) provides a balance between preserving essential geometries and avoiding unnecessary ones, thereby achieving optimal completion accuracy (the lowest $l1$-err. and CD).

\begin{minipage}{.46\textwidth}
\flushleft
\makeatletter\def\@captype{table}
\makeatother\caption{Choice of occupancy threshold. Extremely low or high values yield worse results.}
\label{tab:supp_threshold}
\renewcommand\tabcolsep{8pt}
\vspace*{2.1mm}
\resizebox{1.\textwidth}{!}{
\begin{tabular}{l| c c c }
\toprule
Threshold~$\tau$ & $l_1$-err.~$\downarrow$ & CD~$\downarrow$ & IoU~$\uparrow$ \\
\midrule
1 & 0.08 & 4.21 & 66.6 \\
2 & 0.05 & 4.09 & 68.2 \\
3~\small{(Ours)} & \textbf{0.05} & \textbf{3.97} & 68.3 \\
4 & 0.06 & 4.10 & \textbf{68.4} \\
5 & 0.07 & 4.13 & 67.8 \\
\bottomrule
\end{tabular}
}
\end{minipage}
\hfill
\begin{minipage}{.5\textwidth}
\flushright
\makeatletter\def\@captype{figure}
\vspace*{2mm}
\makeatother\caption{Accuracy (MMD) and diversity (TMD) curves with varying training iterations.}
\vspace*{0.8mm}
\includegraphics[width=.92\columnwidth]{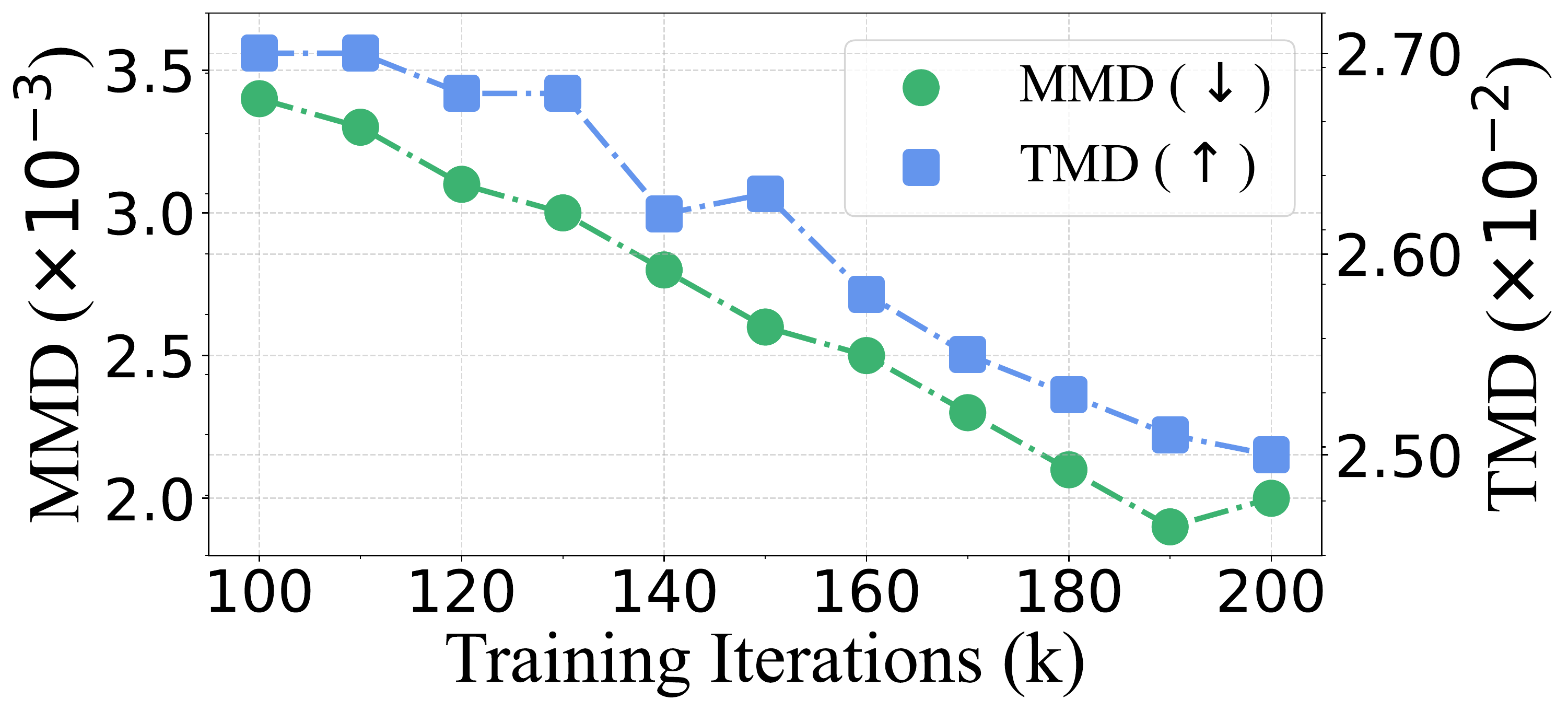}
\label{fig:supp_tmd_mmd_iter}
\end{minipage}

\begin{table}[htbp]
\caption{Finetuning on eight unseen object categories with limited data. Our model shows substantial improvements with incremental data. The 0\% and 100\% indicate zero and full data usage, respectively.}
    \centering
    \vspace*{2mm}
    \renewcommand\tabcolsep{8pt}
     \resizebox{\textwidth}{!}{
     \begin{tabular}{l 
 | c | c c c c c c c c }
\toprule 
Data Ratio & Avg. CD~$\downarrow$ & Bag & Lamp & Bathtub & Bed & Basket & Printer & Laptop & Bench \\  
\specialrule{0em}{2pt}{0pt}
\hline
\specialrule{0em}{2pt}{0pt}
0\% & 3.19 & 2.98 & 3.54 & 2.87 & 3.24 & 3.70 & 3.46 & 2.85 & 2.87 \\
1\% & 2.84 & 2.54 & 3.27 & 2.46 & 2.98 & 3.24 & 3.25 & 2.49 & 2.52 \\
5\% & 1.84 & 1.67 & 1.93 & 1.63 & 2.01 & 1.96 & 2.04 & 1.81 & 1.70\\
10\% & 1.34 & 1.23 & 1.43 & 1.20 & 1.40 & 1.49 & 1.48 & 1.28 & 1.19 \\
100\% & 1.02 & 0.98 & 1.07 & 0.95 & 1.05 & 1.12 & 1.10 & 1.00 & 0.95 \\
\bottomrule
\end{tabular}
}
\label{tab:supp_data_efficient}
\end{table}

\subsection{Impact of Training Iteration}

Fig.~\ref{fig:supp_tmd_mmd_iter} plots the performance of our model over a range of training iterations from 100k to 200k. 
As the training iterations increase, the shape completion accuracy improves (with lower MMD) while the completion diversity gradually decreases (with lower TMD).
This trend reveals that, given more training time, the model learns to better fit the target distribution.

\subsection{Data-efficient Finetuning on Unseen Categories}
\label{subsec:supp_data_efficient}

In Table~\ref{tab:supp_data_efficient}, we evaluate the model's ability to complete ShapeNet objects~\cite{chang2015shapenet} of unknown categories when finetuned with limited data. 
To this end, we first divide the data from unseen classes into a 7:3 train-test split. Then we finetune our model using varying proportions of the training set (1\%, 5\%, and 10\%).
Here, a ratio of 0\% indicates no finetuning process, following the setting in Table.~\textcolor{red}{2}, while 100\% means using the entire training set. A lower CD denotes better completion accuracy.

With just 1\% finetuning data, the average CD decreases by 10.9\% (from 3.19 to 2.84).
A more substantial improvement is observed when the data ratio increases to 5\%, with a nearly 1 point decrease in average CD compared to the 1\% ratio. 
The trend of improvement continues for a 10\% data ratio, where the model impressively approaches the performance achieved using the full training set. 
These results demonstrate that our model has a robust few-shot learning capability and can generalize well from a small amount of out-of-distribution data.

\subsection{Failure Cases}
\label{subsec:supp_failure}
\begin{figure}[htbp]
	\begin{center}
\includegraphics[width=1.0\columnwidth]{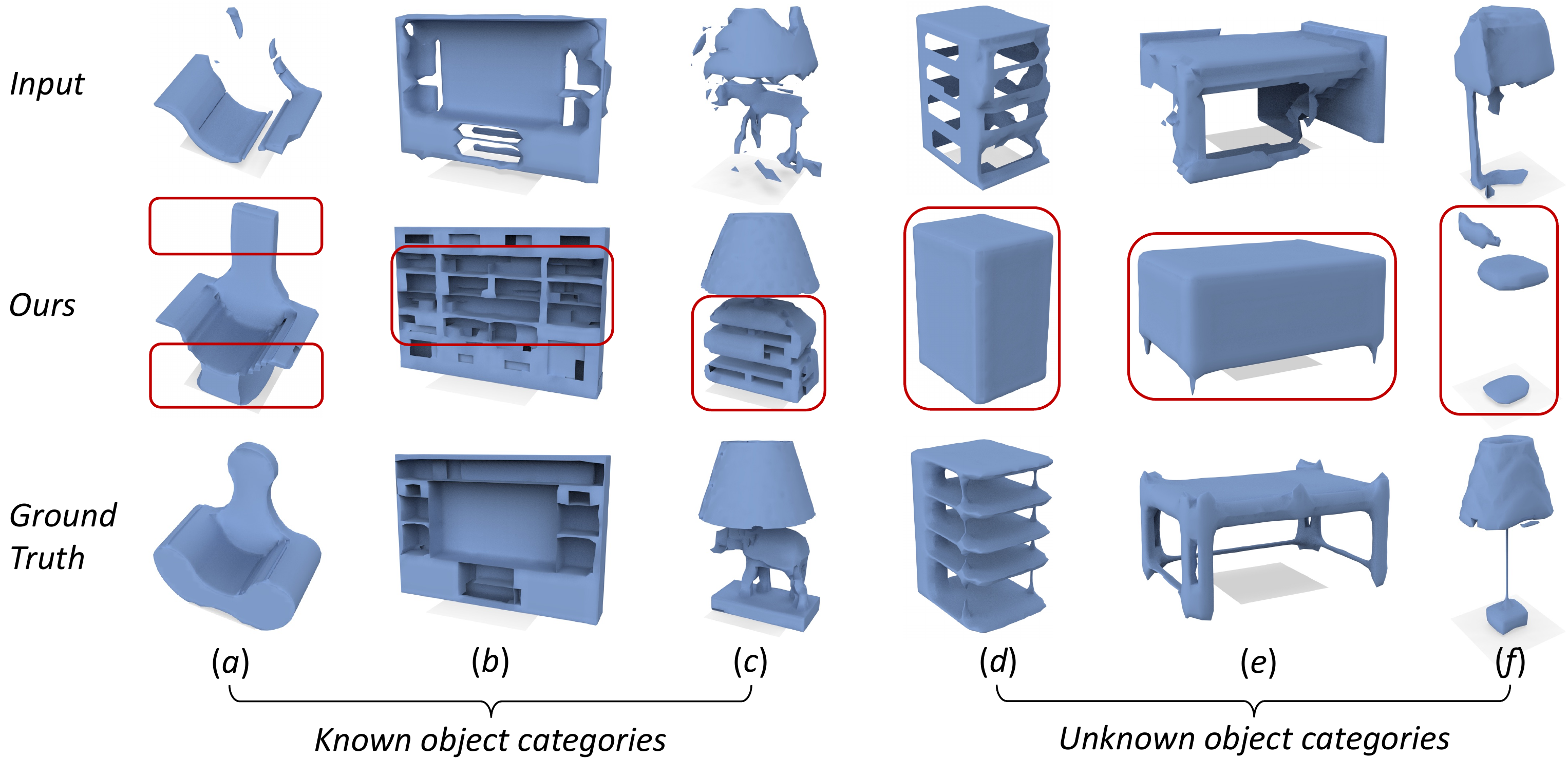}
	\end{center}
	\caption{Failure cases on known and unknown object categories. Our model may produce inaccurate or improbable completions when faced with overly sparse inputs (a), atypical shapes (b-c), complex structures (d), and high noise levels (e-f). The red boxes highlight the difference with ground truths.}
	\label{fig:supp_failure}
\end{figure}

Fig.~\ref{fig:supp_failure} showcases certain failure instances in our completion results. 
For shape completion on known object categories (a-c), given the overly sparse input, our model struggles with producing a shape that aligns with the ground truth (see (a) and (b)).
In example (c), the model fails to complete a non-standard structure, such as an elephant beneath a lamp. This can be attributed to the model's tendency to generate shapes based on frequently seen patterns during training, while the elephant belongs to an atypical structure in lamp categories.
For unknown object categories (d-f), the model faces additional challenges. The case (d) reveals that our model may favor simple structures when the input shape is too complex.
Cases (e) and (f) further show the model's difficulty in handling substantial input noise, which results in inaccurate or improbable completions.

\section{Quantitative Visualizations}
\label{sec:supp_vis}

\begin{figure}[htbp]
	\begin{center}
\includegraphics[width=1.0\columnwidth]{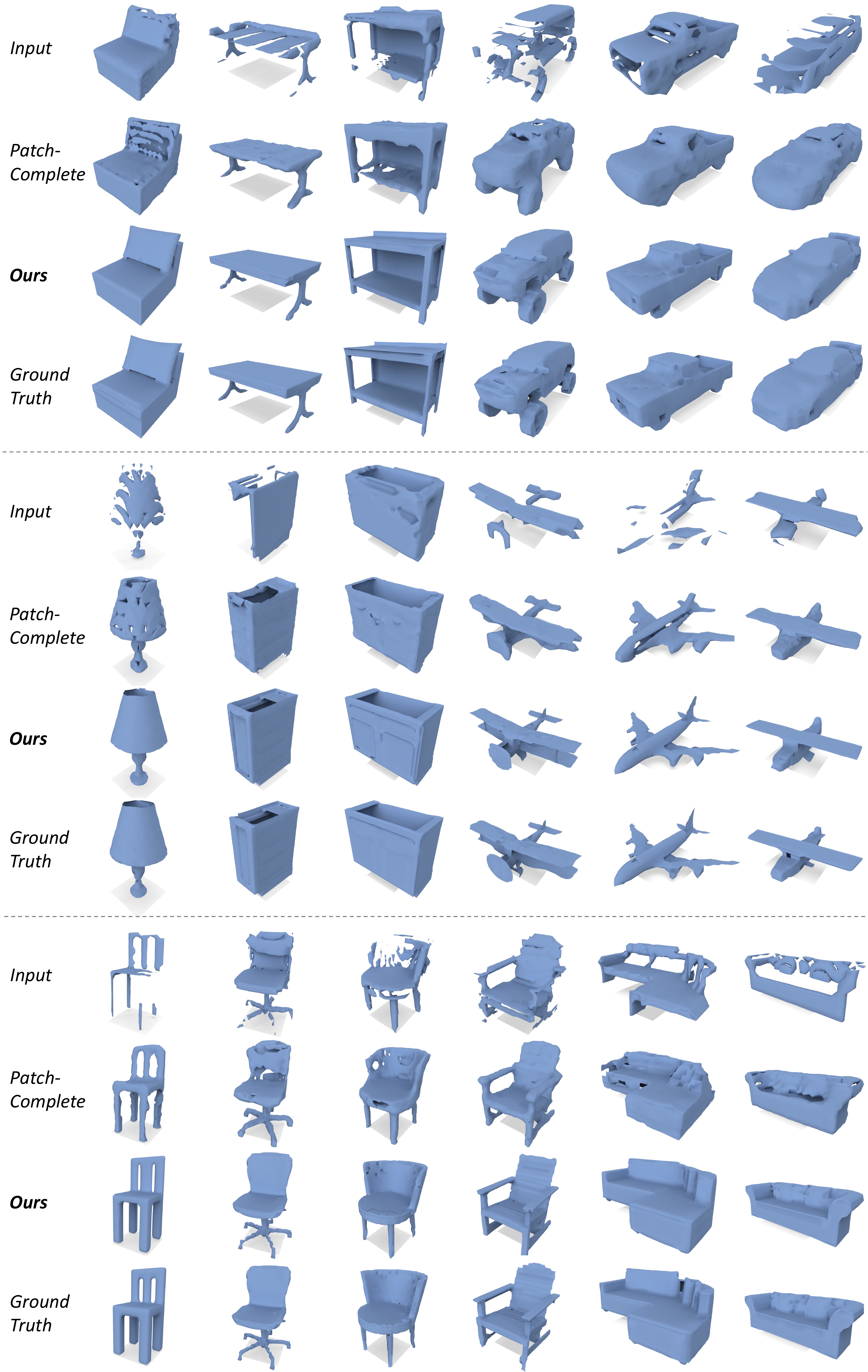}
	\end{center}
	\caption{Quantitative results on completing objects of diverse known categories. Our method significantly outperforms SOTA PatchComplete~\cite{rao2022patchcomplete} on both the completion quality and accuracy.}
	\label{fig:supp_known}
\end{figure}

\textbf{Visualizations on known object categories.} Fig.~\ref{fig:supp_known} shows quantitative results on diverse object categories produced
by SOTA PatchComplete and our \ourMethod. Our method produces the completion results with much fewer artifacts and more realistic shapes.
Our completions are also highly accurate, closely recovering the ground-truth shapes.

\begin{figure}[htbp]
	\begin{center}
\includegraphics[width=1.0\columnwidth]{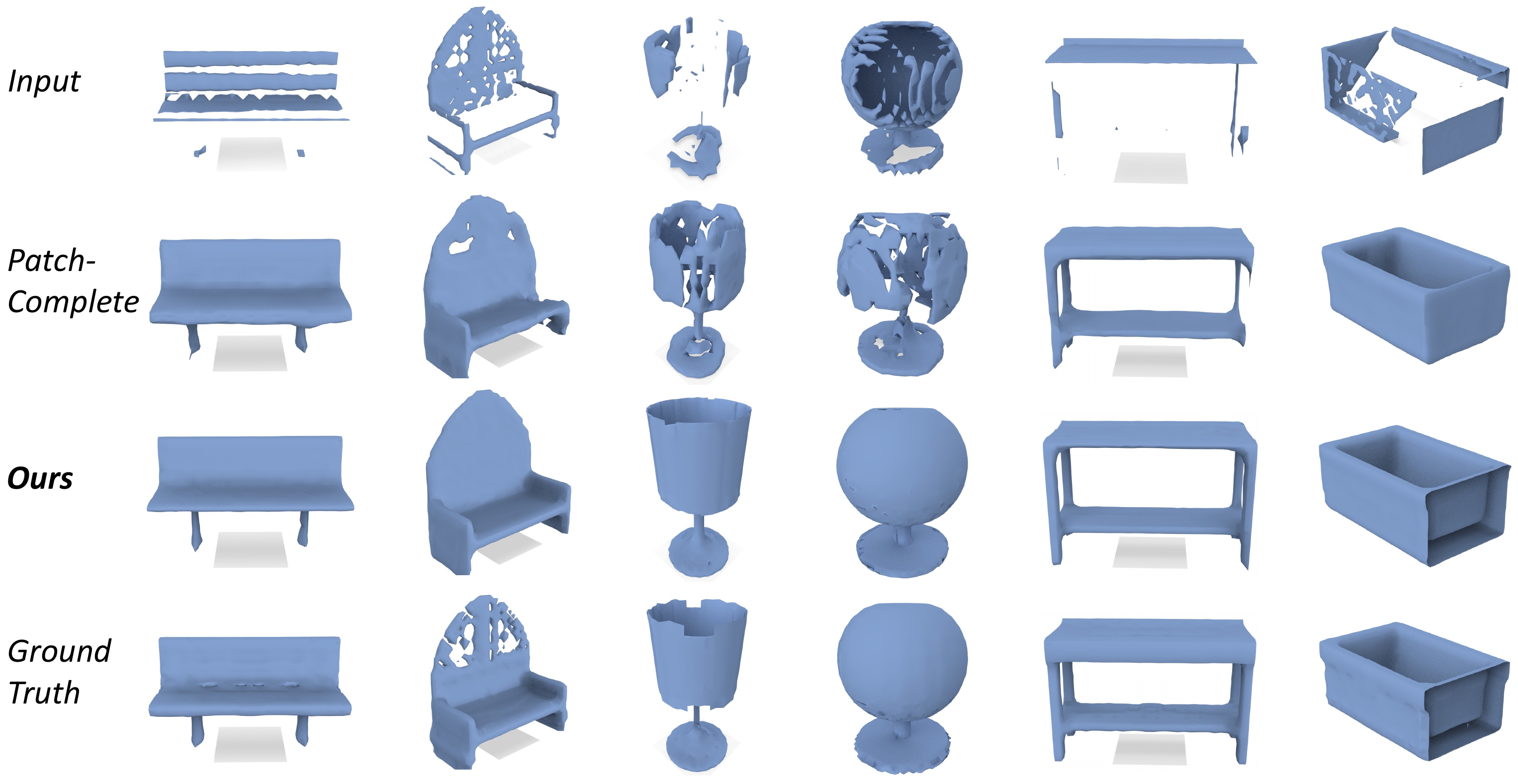}
	\end{center}
	\caption{Quantitative results on completing objects of entirely unseen categories. Our significantly outperforms SOTA PatchComplete~\cite{rao2022patchcomplete} on both the completion quality and accuracy.}
	\label{fig:supp_unknown}
\end{figure}

\textbf{Visualizations on unseen object categories.} As shown in Fig.~\ref{fig:supp_unknown}, across a diverse set of entirely unseen object categories, our method also achieves superior completion results over PatchComplete, preserving better global coherence and local details. Note that we do not employ any zero-shot designs while PatchComplete does.

\begin{figure}[htbp]
	\begin{center}
\includegraphics[width=1.0\columnwidth]{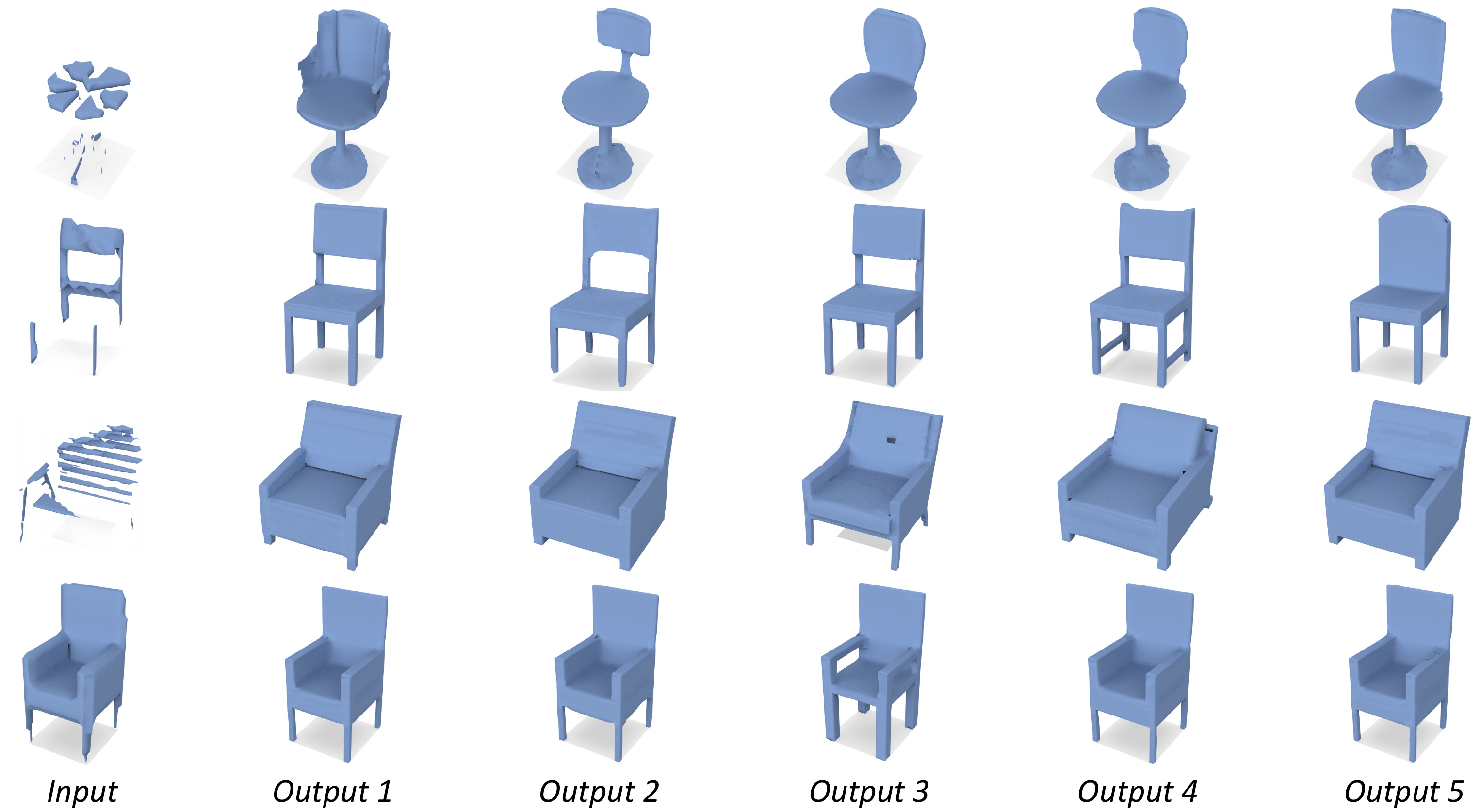}
	\end{center}
	\caption{Multimodal completion results on ShapeNet Chair class. We run the model five times for the same input. The level of input sparsity affects the diversity and certainty of the shape completion.}
	\label{fig:supp_multimodal}
\end{figure}

\textbf{Multimodal completion results.} In Fig.~\ref{fig:supp_multimodal}, we show multiple plausible completion results produced by our model from the same partial shape input.
For sparser input shapes, our model can explore different possibilities to fill in the missing regions and yield more diverse results (\eg, the first row). In contrast, for the inputs with higher completeness levels, our control-based design ensures the model to output more consistent 3D shapes (\eg, the last row).

\begin{figure}[htbp]
	\begin{center}
\includegraphics[width=1.0\columnwidth]{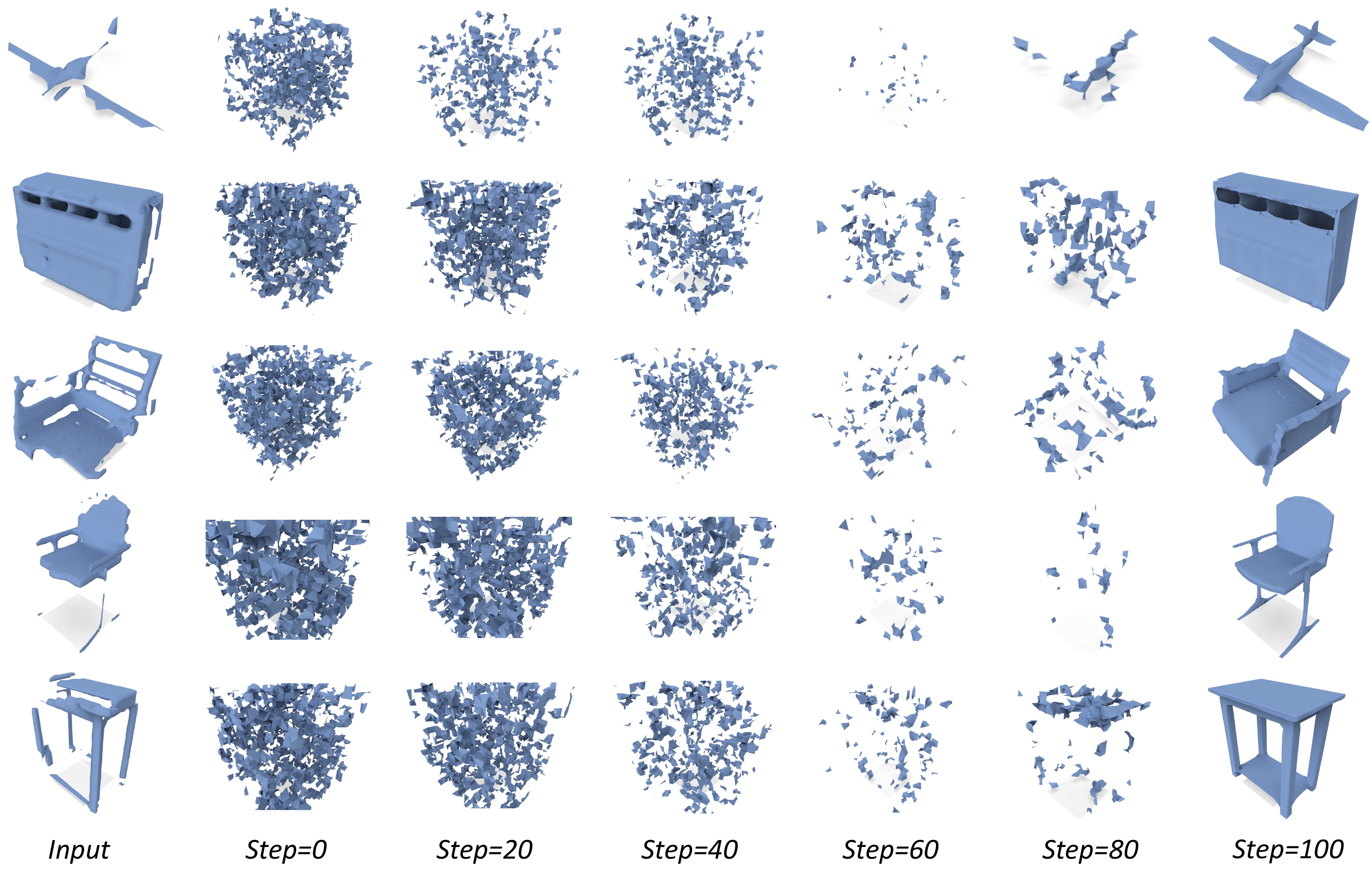}
	\end{center}
	\caption{The denoising process that gradually converts the noises to completed shapes (from left to right). We visualize the produced shapes at varying time steps (0, 20, 40, 60, 80, and 100).}
	\label{fig:supp_step}
\end{figure}

\textbf{Visualizations of denoising process.} In Fig.~\ref{fig:supp_step}, we visualize the produced shapes at different time steps during the inference stage. Our model progressively converts the noises into clean 3D shapes.

\section{Limitations and Potential Solutions}
\label{sec:supp_limitation}

First, our model may struggle to complete highly irregular or noisy shapes, as extensively discussed in Sec.~\ref{subsec:supp_failure}.
Yet, our model's multimodal capacity will increase the likelihood of producing satisfactory results, enabling it to tackle this problem more effectively than deterministic methods. 
With more diverse training data, the model's performance on completing these hard shapes could be improved.

Like most diffusion models~\cite{ho2020denoising,nichol2021improved}, another limitation of \ourMethod~is the substantial computational requirements due to the iterative completion process. Despite employing the technique in work~\cite{song2020denoising} to reduce sampling steps by ten times, we still need 100 steps to achieve competitive results, which costs around 3~4 seconds per shape on an RTX 3090 GPU. The extended test time could cap its potential for real-time or resource-constrained applications. Future work will leverage the advances in fast sampling techniques (\eg,~\cite{zheng2022fast}) to accelerate inference speed.

Also, the dense 3D CNN architecture in our implementation limits the model ability to handle high-resolution 3D shapes, due to the cubic increase in computational costs with volume size. A potential solution could be replacing dense CNNs with efficient 3D network modules, such as SparseConv~\cite{3DSemanticSegmentationWithSubmanifoldSparseConvNet} or Octree-based layers~\cite{wang2017cnn}, while remaining other essential designs of our framework.

At last, although our model shows robust generalizability to unseen object classes, its performance may be adversely affected by the quality and diversity of the training data. 
In cases where object classes or shapes deviate significantly from the training set, the model may underperform.
Therefore, careful selection of training data is needed to boost completion robustness.

In conclusion, despite the current limitations, they also present opportunities for model improvement. By addressing these issues, we believe the full potential of our model can be further realized.

\section{Broader Impact}
\label{sec:supp_impact}

On the positive side, the potential applications of our work are widespread. \ourMethod~could contribute to fields such as computer vision, robotics, virtual reality, and many others. 
For instance, in computer vision and robotics, our method can significantly enhance object reconstruction capabilities, providing more accurate and realistic models that facilitate object recognition, manipulation, and robot navigation. Similarly, in virtual reality or 3D printing, our model is able to complete or refine 3D models, enriching the user experience and the quality of end products.

Moreover, our model provides a flexible balance between the completion diversity and accuracy. This attribute can be tailored to suit various application needs, thereby broadening its potential usability across different tasks.

On the other hand, it is crucial to consider potential negative implications.
As with any AI technology, there are risks associated with misuse. For instance, if used for recreating personal items without consent, it could lead to unwarranted privacy intrusions. In addition, the automation facilitated by our model may also displace jobs involving manual 3D modeling or shape completion.

To conclude, while our research holds promising potential, it is essential to responsibly manage its broader impacts. We advocate for developing this technology in a way that maximizes societal benefits and minimizes potential negative effects.

\end{document}